\documentclass[10pt,twocolumn,letterpaper]{article}

\usepackage[pagenumbers]{cvpr} 

\usepackage{graphicx}
\usepackage{amsmath}
\usepackage{amssymb}
\usepackage{booktabs} 
\usepackage[utf8]{inputenc}
\usepackage{multirow}
\usepackage{placeins}
\usepackage{soul}
\usepackage{subcaption}
\usepackage{todonotes}
\usepackage{caption}
\usepackage{relsize}

\usepackage[pagebackref,breaklinks,colorlinks]{hyperref}

\usepackage[capitalize]{cleveref}
\crefname{section}{Sec.}{Secs.}
\Crefname{section}{Section}{Sections}
\Crefname{table}{Table}{Tables}
\crefname{table}{Tab.}{Tabs.}

\def\cvprPaperID{*****} 
\def\confName{CVPR}
\def\confYear{2022}

\begin{document}

\title{Improving the Robustness of Quantized Deep Neural Networks to White-Box Attacks using Stochastic Quantization and Information-Theoretic Ensemble Training}

\author{Saurabh Farkya \\
Center for Vision Technologies \\
SRI International \\
{\tt\small saurabh.farkya@sri.com}
\and
Aswin Raghavan \\
Center for Vision Technologies \\
SRI International \\
{\tt\small aswin.raghavan@sri.com}
\and
Avi Ziskind \\
Center for Vision Technologies \\
SRI International \\
{\tt\small avi.ziskind@gmail.com}
}
\maketitle

\begin{abstract}
Most real-world applications that employ deep neural networks (DNNs) quantize them to low precision to reduce the compute needs. 
We present a method to improve the robustness of quantized DNNs to white-box adversarial attacks.
We first tackle the limitation of deterministic quantization to fixed ``bins'' by introducing a differentiable Stochastic Quantizer (SQ).
We explore the hypothesis that different quantizations may collectively be more robust than each quantized DNN.
We formulate a training objective to encourage different quantized DNNs to learn different representations of the input image.
The training objective captures diversity and accuracy via mutual information between ensemble members. Through experimentation, we demonstrate substantial improvement in robustness against $L_\infty$ attacks even if the attacker is allowed to backpropagate through SQ 
(e.g., > 50\% accuracy to PGD(5/255) on CIFAR10 without adversarial training), 
compared to vanilla DNNs as well as existing ensembles of quantized DNNs. We extend the method to detect attacks and generate robustness profiles in the adversarial information plane (AIP), towards a unified analysis of different threat models by correlating the MI and accuracy. 
\end{abstract}

\section{Introduction}
\label{sec:intro}

In the context of deep image classifiers, $f_{\theta}: X \rightarrow \Delta(Y)$, where $X$ is an image and $\theta$ denotes the neural network parameters, an adversarial attack is a perturbation $\delta$ that is optimized such that $f_{\theta}(x+\delta) \neq y^*$ with respect to the true label $y^{*}$, with a constraint on the size of the attack e.g.\!  $\left\Vert \delta \right\Vert _{\infty}$ is small. 

\begin{figure}
    \centering
    \includegraphics[width=0.95\linewidth]{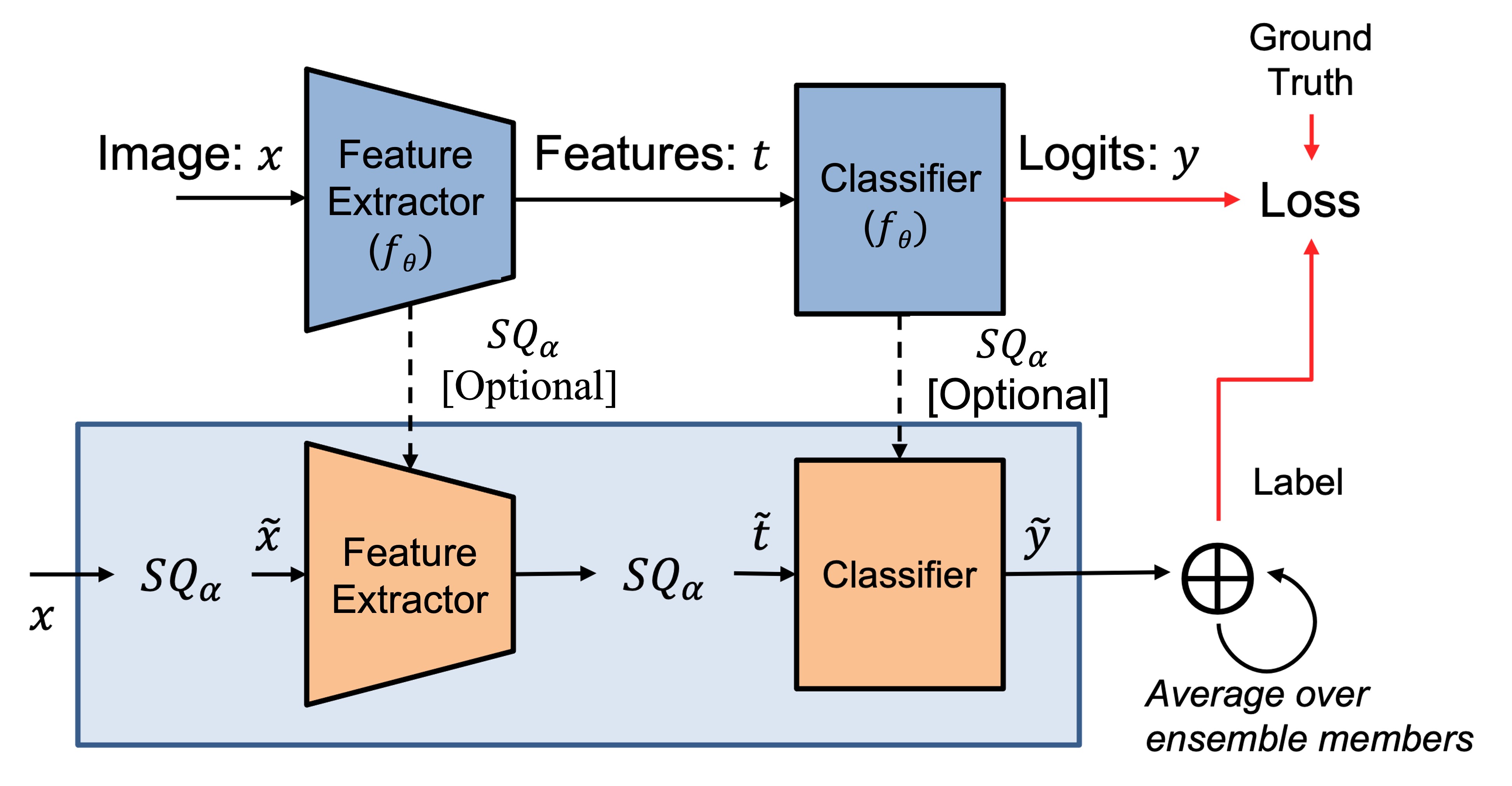}
    \caption{Overview of our Stochastic Quantization (SQ)-based ensemble. Our approach hardens the pre-trained feature extractor in blue by injecting SQ at the input and feature layers.}
    \label{fig:rfs_architecture}
\end{figure}

DNNs (especially larger models) deployed to real-world systems tend to use quantization to reduce the compute resources needed. Quantization reduces the precision or bit-width of weights or activations. Quantized DNNs have limited expressivity, determined by the nature of the quantization function and the number of bits of precision. Considering a white-box threat model, wherein the attacker has complete knowledge of the quantization function, the quantized version of $f_\theta$ can be more vulnerable than $f_\theta$. Significant advancements have been made to improve the adversarial robustness of $f_\theta$, but may not help lead to a robust quantized DNN. 

We first tackle the limitation of deterministic quantization to fixed ``bins'' (e.g., 8-bit unsigned integers) as it is a fundamental challenge to achieve robust quantized DNNs. We introduce a novel Stochastic Quantizer (SQ) (Section \ref{sec:sq}) that induces a continuous, sparse, and low-variance probability distribution over bins. Our SQ works for any target bit-precision and is more general than previous binary SQ  \cite{courbariaux2015binaryconnect}. 
Repeated sampling from SQ generates our ensemble of (any number of) DNNs without incurring a training computational complexity that scales with the number of members. The hypothesis is that different quantizations may collectively be more robust compared to each quantized DNN. 


One basis for this hypothesis is in \cite{sen2020empir} where quantized DNNs of different precisions were combined \emph{as is} (without fine-tuning in concert) and showed promising robustness against adversarial attacks \cite{sen2020empir}. 
We explore further and formulate a training objective to encourage different quantized DNNs to learn different representations of the input image. We train the ensemble to maximize the conditional Shannon entropy $H(\tilde{T}|\tilde{X})$, where $\tilde{T}$ is a random variable denoting the deep quantized representation (see Figure \ref{fig:rfs_architecture}). Equivalently, we add the mutual information (MI) $I(\tilde{X};\tilde{T})=H(\tilde{T})-H(\tilde{T}|\tilde{X})$  as a regularizer to the usual cross-entropy classification loss. A key advantage of the quantization-based approach is that direct MI calculation is tractable because it is computed over discrete random variables with known probability mass functions (PMFs) induced by SQ (Section \ref{sec:mi_stuff}).


Note that the MI in our method can be calculated from a single image and without distributional assumptions on the training data because the PMF is induced by random noise internal to the DNN. We find that regularization (of $\theta$) is needed to avoid unbounded amplification of this noise. We regularize the average spacing between the bins in the representation space given stochastically quantized inputs. Further research is needed to make precise the connection to the interpretation of the Lipschitz constant in quantized DNNs \cite{finlay2018lipschitz,lin2019defensive}. 




We show empirical results for end-to-end adversarial attacks over a large number of experiments with varying SQ noise ($\alpha$), MI regularization ($\beta$), and ``Lipschitz'' regularization ($\mu$). We show significant gains in robustness compared to a pre-trained vanilla DNN,  
ensemble of vanilla DNNs trained from different random initializations \cite{pang2019improving}, 
a quantized DNN trained with Lipschitz regularization \cite{lin2019defensive}, and 
\cite{sen2020empir}'s ensemble of pre-trained quantized DNNs.

Beyond the experiments showing accuracy under perturbations, we visualize the change in MI values against the change in accuracy for attacks of different types and strengths. This visualization called the Adversarial Information Plane (AIP), may enable extrapolation and a unified understanding of different threat models in terms of MI. 
The AIP from our experiments identified several relative attack hardness properties that vary by attack, dataset, and NN architecture. 
Finally, we leverage the estimated MI for attack detection by comparing it with the average MI over unperturbed data, allowing the model to abstain from making predictions on noisy data. 

\section{Methodology}
\subsection{Stochastic Quantization}
\label{sec:sq}
\begin{figure*}
    \centering
    \begin{subfigure}{0.3\textwidth}
        \includegraphics[width=\textwidth, height=1in]{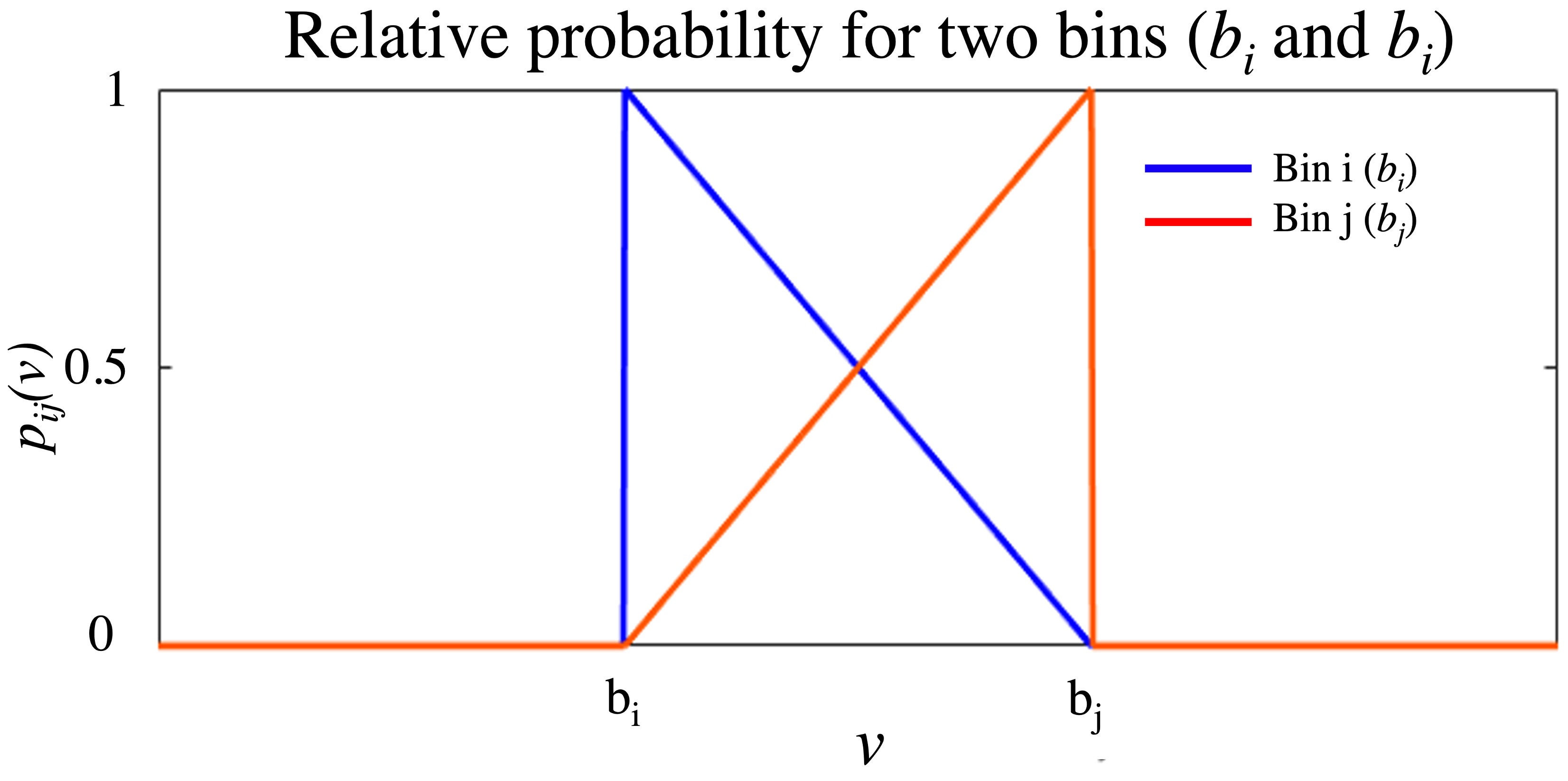}
        \label{fig:SQ1}
    \end{subfigure}
    \begin{subfigure}{0.3\textwidth}
        \includegraphics[width=\textwidth, height=1in]{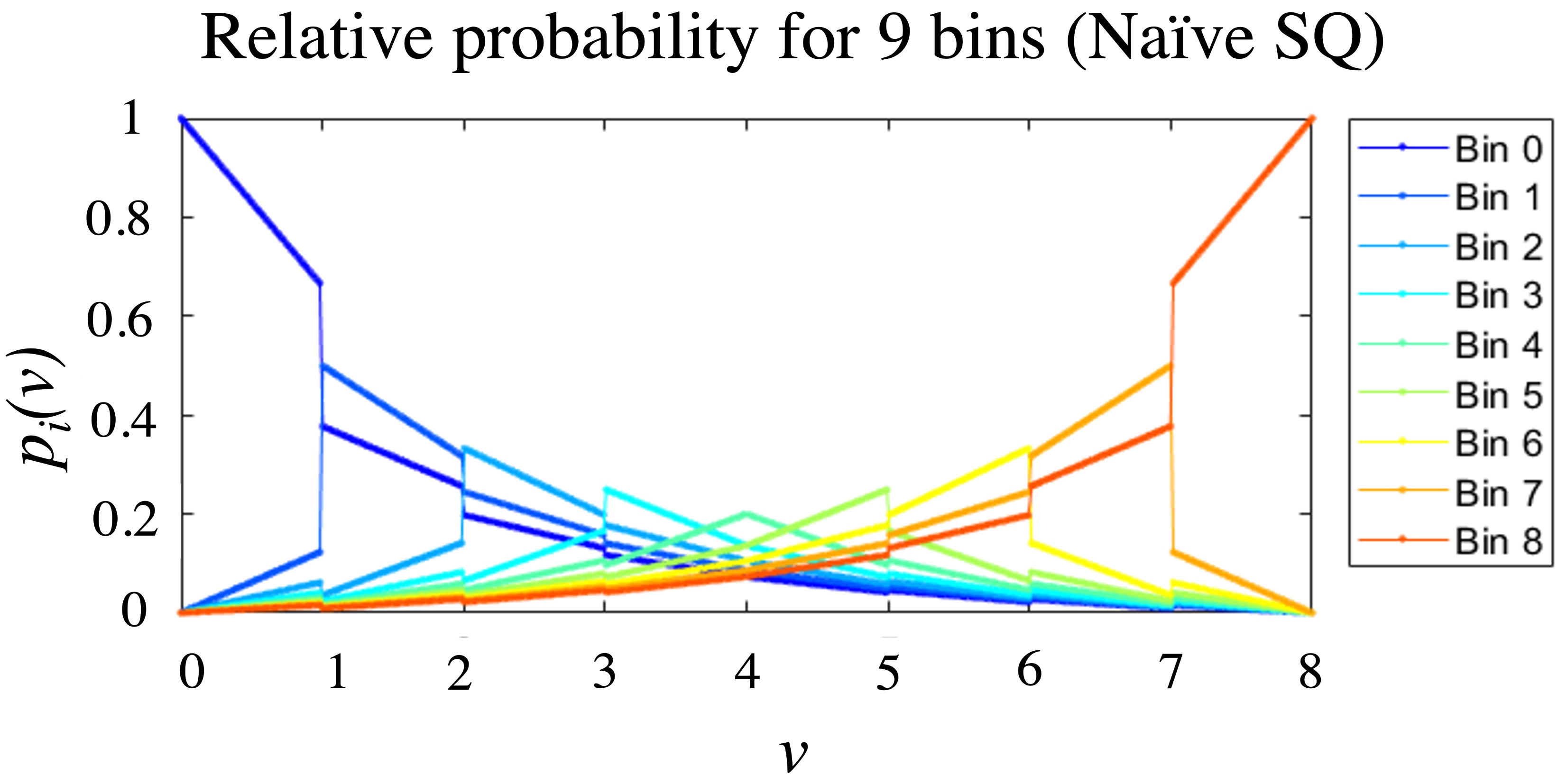}
        \label{fig:SQ_naive_pi}
    \end{subfigure}
    \begin{subfigure}{0.3\textwidth}
        \includegraphics[width=\textwidth, height=1in]{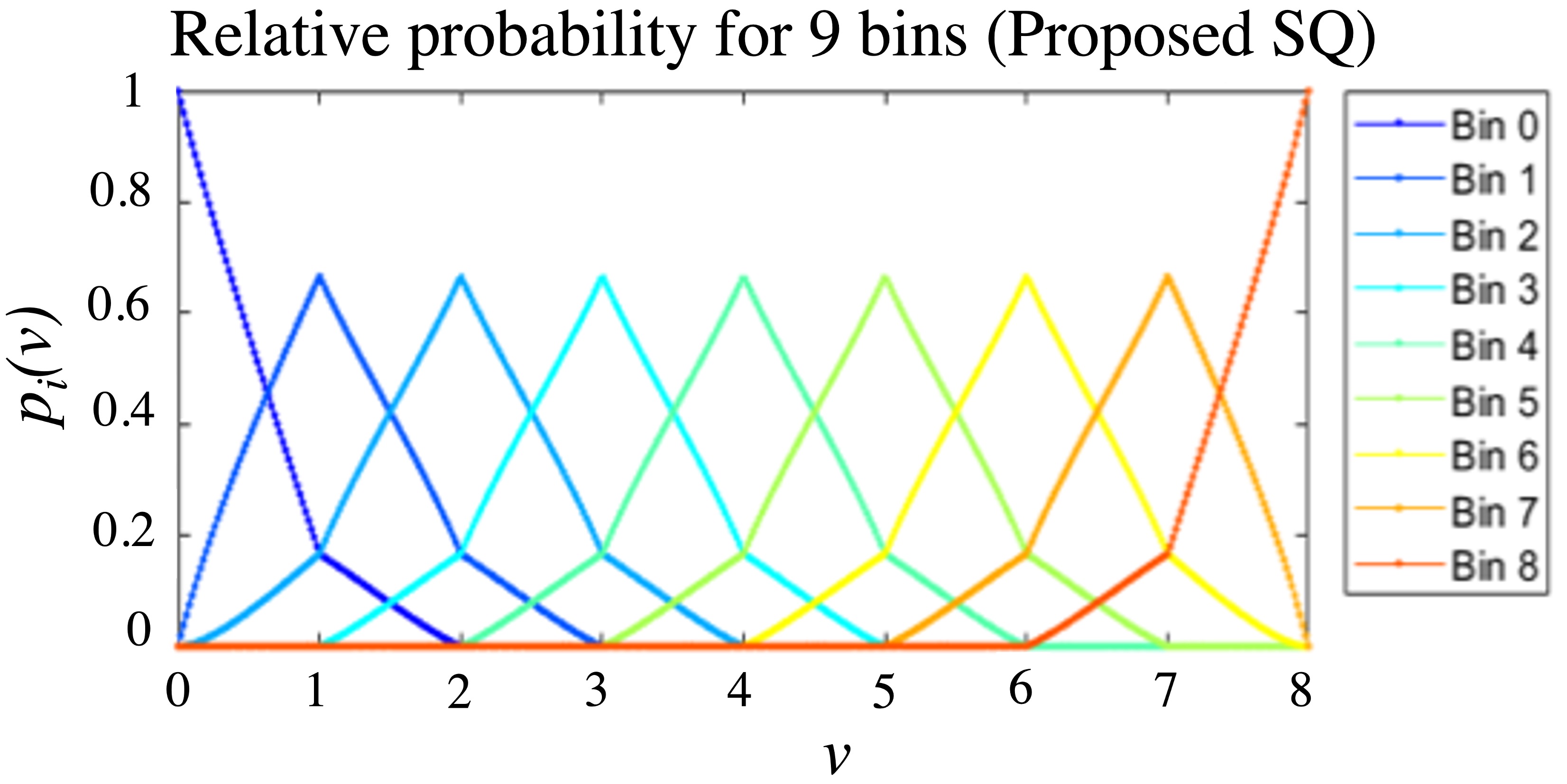}
        \label{fig:SQ_ours}
    \end{subfigure}
    \caption{(a) Bin probabilities for SQ to two bins \cite{courbariaux2015binaryconnect} (Eq. \ref{eq:SQ1}). (b) Unbiased SQ for nine bins by extending Eq. \ref{eq:SQ1}. (c) Our SQ (Eq. \ref{eq:SQ_ours} with $\alpha=2$)}
    \label{fig:combined}
\end{figure*}

Quantization \cite{gholami2021survey, guo2018survey, yang2019quantization} is widely used for accelerating inference and training of deep neural networks (DNNs). 
A (scalar) quantization function $q: \mathbb{R} \rightarrow \mathcal{B}$ maps a real number to an element of an ordered set of values known as $n$ ``bins'', $\mathcal{B} = \{b_i\}, i=0,\ldots,n$, where each $b_i \in \mathbb{R}$ and $b_{i+1} > b_i$. The bins can have restricted support like  binary \cite{courbariaux2015binaryconnect}, ternary \cite{li2016ternary}, powers-of-two \cite{parajuli2018generalized}, or integers \cite{banner2018scalable}. In this work, we use real-valued bins evenly spaced with $\delta = b_{i+1} - b_i, \forall i$, which can be mapped to integer bins in post-processing. The values of $b_0$ and $b_n$ can be fixed or based on the minimum and maximum values in a given image or feature vector. 

In Stochastic Quantization (SQ) \cite{courbariaux2015binaryconnect}, the quantized value is a random variable, $q(v) \sim p(v)$, with a categorical distribution $p: \mathbb{R} \rightarrow \Delta(\mathcal{B})$ supported on the bins. Differentiable sampling from $p(v)$ allows training with standard gradient methods.

 
An SQ is unbiased if $E_{p(v)}[q(v)] = v$.  
Consider the binary SQ scheme of \cite{courbariaux2015binaryconnect} defined for values $v \in [b_0, b_1]$ given by


\begin{equation}
    p(v) = \begin{cases} 
    b_0 & \text{w.p. } \frac{b_1 - v}{b_1 - b_0} \\
    b_1 & \text{w.p. } \frac{v - b_0}{b_1 - b_0} 
    \end{cases}.
    \label{eq:SQ1}
\end{equation}
where w.p. denotes with probability. With this distribution (Figure~\ref{fig:combined}(a)), $q(v)$ is unbiased for $v \in [b_0, b_1]$.
One can extend this scheme to an arbitrary number of bins $k$ as follows. Let $p_{ij}(v)$ denote the quantization probabilities for two bins $b_i < b_j$ and $p_i(v)$ denote the probability of mapping $v$ to $b_i$, $p_i(v) = \frac{1}{Z}\tilde{p}_i(v)$,  $\tilde{p}_i(v) = \sum_{j} p_{ij}(v)$, where the summation is over valid $j$ such that $v \in [b_i, b_j]$, and 
where the normalization constant $Z = \sum_{i} \tilde{p}_i(v)$. While this scheme is unbiased, it leads to discontinuities at bin boundaries (Figure~\ref{fig:combined}(b)), and moreover exhibits high sample variance and low sparsity caused by non-zero probabilities assigned to all bins.

To address these problems, we introduce a parameter \textbf{$\alpha$} that controls the sparsity and variance of SQ by limiting non-zero probabilities to bins that are at most $\alpha$ bins away from $v$.
We scale the bin probabilities as a function of distance normalized by the spacing between the bins. 
Let $\delta_i(v) = \frac{|v-b_i|}{\delta}$ be the normalized distance in a number of bins. The probabilities of $b_i$ and $b_j$ are weighted by $ReLU(1-\frac{\delta_i}{\alpha})$, where $ReLU(t)=max(0,t)$. The quantization distributions are given by,
\begin{equation}
p_{ij}(v;\alpha) = \begin{cases}
        b_i \text{ w.p. } \frac{b_j-x}{b_j-b_i}\rho(1 - \frac{\delta_i}{\alpha})\rho(1- \frac{\delta_j}{\alpha}) \\
        b_j \text{ w.p. } \frac{x-b_i}{b_j-b_i}\rho(1 - \frac{\delta_i}{\alpha})\rho(1- \frac{\delta_j}{\alpha}) \\
     \end{cases}
     \label{eq:SQ_ours}
\end{equation}
where $\delta_k = \frac{|v-b_k|}{\delta}$ and $\delta = b_2 - b_1$ is the bin spacing. 

This formulation results in probabilities free of discontinuities (Figure~\ref{fig:combined}(c)), with reduced variance compared to the na\"ive approach due to distant bins having less influence, with zero probability assigned to bins that are more than $\alpha$ bins away. The quantization noise is bounded by $\alpha \delta$, $\left\Vert \mathbf{v} - \tilde{v} \right\Vert \leq \mathcal{O}(\alpha \delta)$, and for an image $\mathbf{X}$, $\|\mathbf{X} - \tilde{x}\|_{\infty} \leq \mathcal{O}(\alpha\delta)$. Empirically, we found the bias to be low. The probabilities are piecewise differentiable with respect to $\alpha$ and $v$ using the Gumbel-Softmax reparameterization \cite{jang2016categorical} for categorical distributions, however, we set the value of $\alpha$ as a hyperparameter in this paper.  
The sparsity due to $\alpha$ allows for an efficient implementation that reduces the computation from  $O(|\mathcal{B}|^2)$ bins with non-zero probability to $O(\alpha^2)$.

\subsection{Ensemble Training}
\label{sec:mi_stuff}

We formulate the ensemble learning problem as training a DNN $f_\theta$ such that a diverse and performant ensemble is generated when SQ is applied to one or more DNN layers. Figure~\ref{fig:rfs_architecture} shows an overview of our approach where $f_\theta$ is shown in the top half of the architecture. 
Although our technique can be applied to any DNN architecture, we make a simplifying assumption that $f_\theta$ is an image classifier with distinct feature extractor and classifier stages. 

\begin{figure}
    \centering
    \includegraphics[width=0.9\columnwidth, keepaspectratio]{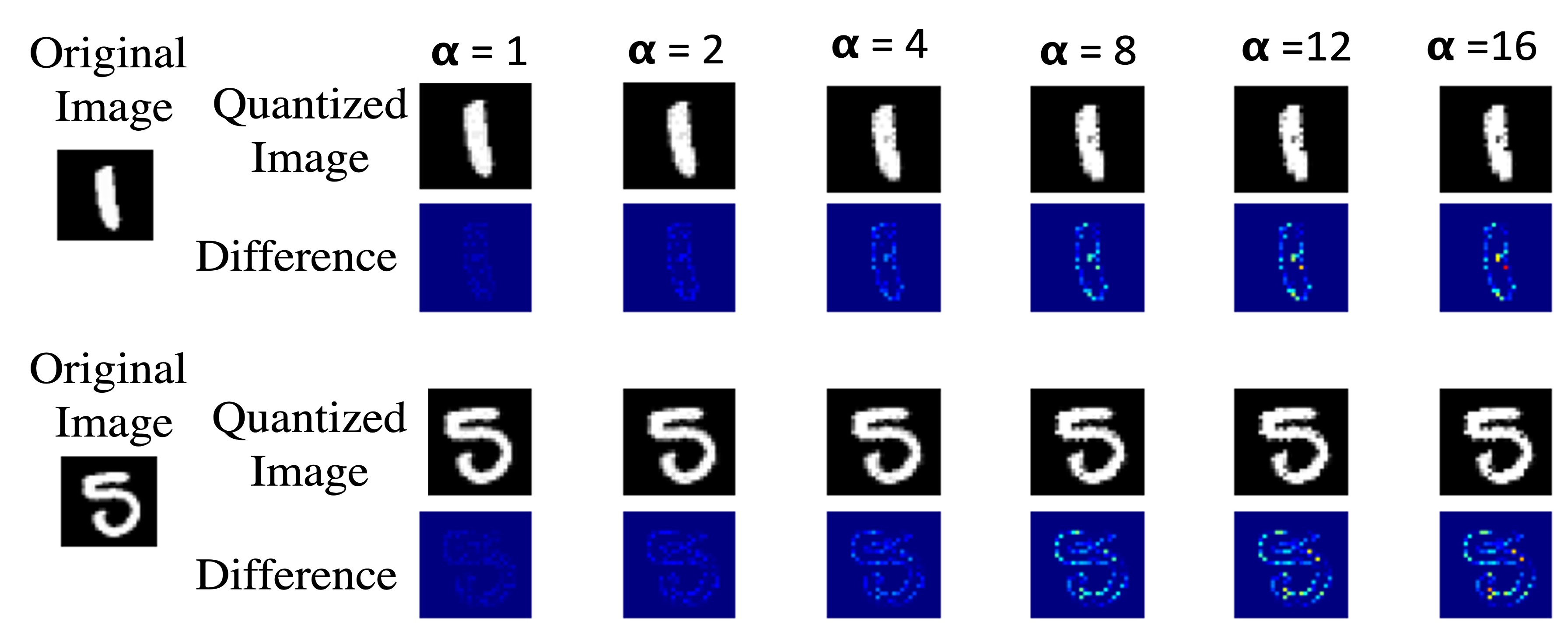}
    \caption{Example perturbations: SQ on MNIST images.}
    \label{fig:SQ_example_image}
\end{figure}

We apply SQ to the input image $x$, 
creating a random variable $\tilde{X}$ with PMF $p(\tilde{X}=x)$ according to Eq.~\ref{eq:SQ_ours}. Each ensemble member forward-propagates a sample $\tilde{x} \sim p(\tilde{X})$. For example, SQ samples for MNIST images are shown in Figure \ref{fig:SQ_example_image}. 
The sample is propagated through the layers of the feature extractor including any quantization steps therein. 
We further apply SQ to the output of the feature extractor. Let $\tilde{T}$ be the random variable resulting from applying SQ to the feature vector with probabilities according to Eq.~\ref{eq:SQ_ours}. Each ensemble member forward-propagates a sample $\tilde{t} \sim p(\tilde{T}|\tilde{X}=\tilde{x})$ through its classifier. Finally, the classifier outputs are aggregated over ensemble members. 

\textbf{Mutual Information Regularization}: We take an information-theoretic approach to the training of a diverse and robust ensemble. We train the ensemble to maximize feature diversity, defined as the Shannon entropy $H(\tilde{T}|\tilde{X})$.   Equivalently, we add the mutual information (MI) $I(\tilde{X};\tilde{T})=H(\tilde{T})-H(\tilde{T}|\tilde{X})$ to the usual cross-entropy loss as a regularizer, 
\begin{equation}
    \min L_{\text{class}}(\theta, \alpha) + \beta I(\tilde{X};\tilde{T}).
    \label{eq:qed_loss}
\end{equation}
where $L_{class}$ is the standard cross-entropy loss used for classification. Computing $I(\tilde{X}; \tilde{T})$ is straightforward as the underlying random variables are discrete with known PMFs. The distribution $p(\tilde{x})$ is the output of SQ applied to one input x, and $p(\tilde{T}|\tilde{x})$  is the output of SQ applied to the feature layer of one sample $\tilde{x}$ i.e. in one ensemble member.

\begin{align}
        H(\tilde{T}_i) &= -\sum_{\tilde{t}_i \in \text{supp}(\tilde{T}_i)} p(\tilde{T}_i)\log{p(\tilde{T}_i)}\\
        H(\tilde{T}_i|\tilde{X}) &= -\sum_{\tilde{x} \sim \tilde{X}} p(\tilde{x}) p(\tilde{T}_i|\tilde{x})\log{p(\tilde{T}|\tilde{x})}\\
        p(\tilde{T}_i) &= \sum_{\tilde{x} \sim \tilde{X}} p(\tilde{T}_i|\tilde{x}) p(\tilde{x})
\end{align}
For multi-dimensional feature vectors, we calculate the MI per feature $\tilde{T}_i$ and average over features. 
In our experiments, we run $n$ forward passes with different $\tilde{x}$. These forward passes can be run concurrently in batches.
We observe our MI estimates to converge quickly (e.g., $n=128$ samples for MNIST with $16$ bins). 
Our method can estimate MI from a single image without any distributional assumptions on the data, in contrast to prior work that used variational approximations \cite{alemi2016deep,pensia2020extracting}, infinite ensembles \cite{shwartz2020information}, or ensembling over batches of images \cite{tishby2000information} to estimate the MI. You can find a worked-out example of MI calculation on a toy DNN in the document\footnote{\label{fn:mi-calculation-doc}\url{https://docs.google.com/document/d/1xNxf2g2tQuqDUIgoLCVaeT44VmeYbnSX/edit?usp=sharing&ouid=116793725997267490520&rtpof=true&sd=true}}.

To access the document, refer to the footnote on page~\pageref{fn:mi-calculation-doc}.

\textbf{Regularization}: Each ensemble member propagates a noisy quantized version of the input image. We know that in some NN architectures, the noise added at the input layer can be amplified by DNN layers and produce very different feature vectors. In our implementation of SQ, the smallest and largest bins are defined as the minimum and maximum values in the feature vector. So the noise amplification results in bins that are farther apart, leading to noisy samples. Recall that the noise added by SQ is bounded by $\alpha \delta$. This issue was mitigated easily by the addition of a regularizer that accounts for the bin spacing, empirically calculated as $b_1^T-b_0^T$ for the feature distribution $\tilde{T}$ averaged over samples $\tilde{x}$. The overall training objective is,

%


 \begin{equation}
         \min L_{\text{class}}(\theta, \alpha) + \beta I(\tilde{X};\tilde{T}) + \mu (b_1^T-b_0^T),
        \label{eq:qed_loss_final}
 \end{equation}
where $\mu > 0, \beta > 0, \alpha > 1$, and all terms have to be averaged over samples $\tilde{x} \sim \tilde{X}$ and a batch dimension $X$, $\tilde{X}=p(X)$ for SQ applied to (Eq. \ref{eq:SQ_ours}) input image $X$, and $\tilde{T}=p(T|\tilde{x})$.

\section{Experiments}

We show empirical results on three image classification datasets --- MNIST \cite{deng2012mnist}, CIFAR10 \cite{Krizhevsky09learningmultiple}, and RESISC45 \cite{Cheng_2017}.  We initialize $\theta$ with pre-trained models (top row of Fig. \ref{fig:rfs_architecture}): LeNet-5 for MNIST \cite{lecun1998gradient} with 99.6\% clean test accuracy, ResNet-18 \cite{He_2016_CVPR} for CIFAR10 and RESISC45 with clean 91.2\% and 85\% clean test accuracy, respectively.

Adversarial robustness is evaluated against white-box attacks FGM \cite{goodfellow2014explaining}, PGD \cite{madry2017towards} with 10-40 iterations, and Square attack \cite{ACFH2020square} of varying strengths.  For each type of attack, we use Expectation over Transformation (EOT) \cite{athalye2018synthesizing} over all ensemble members (i.e.\! samples of SQ). The attacker is able to backpropagate gradients through SQ using exactly the same method used in training (Gumbel-softmax with unchanged hyperparameter). The attacker's objective includes the cross entropy term but does not include the regularization terms in Eq.~\ref{eq:qed_loss_final}.

We evaluated several settings of the hyperparameters in Eq.~\ref{eq:qed_loss_final} for each dataset with a fixed ensemble size of $16$ i.e.\! $16$ samples from SQ during training. We used $16$ bins for the input and feature layer SQ so our models operate in 4 bits of activations only. The 4-bit inference is significantly more efficient than training the attack perturbation at full precision.
Accuracies reported for our method are annotated by the hyperparameters as ($\alpha$/$\mu$/$\beta$/atk.\! samples). $\alpha, \mu, \beta$ refer to the sparsity in SQ, regularization hyperparameter, and MI regularization hyperparameter Eq.~\ref{eq:qed_loss_final}. 

\textbf{Baselines}:
\textbf{Atk.\! samples} refers to the number of samples that the attacker is allowed to draw from SQ. When Atk.\! samples is $16$ (equal to training) we expect lower accuracy than Atk.\! samples = 1 which is a best case baseline. 

\textbf{SQ @ Input}  performs SQ only at the input layer and no SQ at the feature layer, therefore does not perform any regularization. This baseline is similar to randomized smoothing \cite{cohen2019certified} except that it uses SQ to add noise to the input. 

\begin{table*}[ht]
\centering
\smaller[2]
\begin{tabular}{|c|c|c|c|c|c|c|}
\hline
Dataset & Method & Atk.\! samples & Clean & FGM ($\epsilon=0.1 \rightarrow 0.2 \rightarrow 0.3$) & PGD ($\epsilon=0.1 \rightarrow 0.15 \rightarrow 0.3$) \\
\hline
\multirow{4}{1.2cm}{MNIST (LeNet5)} & Vanilla & 1 & 99.02\% & $71.86 \rightarrow 27.32 \rightarrow 15.88$ & $53.2 \rightarrow 19.01 \rightarrow 0.8$ \\   
& SQ @ Input ($\alpha=4$) & 16 & 99.23 & $95.07 \rightarrow 87.16 \rightarrow 76.57$ & $87.67 \rightarrow 75.49 \rightarrow 24.42$ \\
& Ours ($\alpha=4/\mu=1/\beta=0$) & 16 & 99.16 & $95.63 \rightarrow 89.81 \rightarrow 81.83$ & $88.01 \rightarrow 76.4 \rightarrow 34.07$ \\ 
& Ours ($\alpha=4/\mu=1/\beta=0$) & 1 & 98.98 & $95.47 \rightarrow 89.19 \rightarrow 79.33$ & $89.30 \rightarrow 80.13 \rightarrow 44.14$ \\

\hline
 &  & & & FGM ($\epsilon=0.02 \rightarrow 0.04 \rightarrow 0.1$) & PGD ($\epsilon=0.01 \rightarrow 0.02 \rightarrow 0.1$) \\
\hline

\multirow{4}{1.2cm}{CIFAR10 (ResNet18)} & Vanilla & 1 & 91.73\% & $22.57 \rightarrow 15.84 \rightarrow 11.99$ & $22.4 \rightarrow 7.81 \rightarrow 5.54$ \\ 
& SQ @ Input ($\alpha=16$) & 16 & 86.72 & $51.35 \rightarrow 28.12 \rightarrow 12.1$ & $73.72 \rightarrow 48.96 \rightarrow 18.81$ \\
& Ours ($\alpha=16/\mu=1/\beta=10$) & 16 & 85.93 & $55.79 \rightarrow 32.99 \rightarrow 20.12$ & $74.5 \rightarrow 53.13 \rightarrow 24.79$ \\ 
& Ours ($\alpha=16/\mu=1/\beta=10$) & 1 & 78.80 & $69.93 \rightarrow 59.14 \rightarrow 35.36$ & $75.93 \rightarrow 70.1 \rightarrow 54.1$ \\

\hline

 &  & & & FGM ($\epsilon=0.005 \rightarrow 0.01 \rightarrow 0.03$) & PGD ($\epsilon=0.003 \rightarrow 0.005 \rightarrow 0.01$) \\
\hline

\multirow{4}{1.2cm}{RESISC45 (ResNet18)} & Vanilla & 1 & 85.80\% & $48.8 \rightarrow 33.82 \rightarrow 20.7$ & $58.17 \rightarrow 39.97 \rightarrow 20.46$ \\ 
& SQ @ Input ($\alpha=16$) & 16 & 80.51 & $71.97 \rightarrow 55.88 \rightarrow 23.13$ & $76.95 \rightarrow 71.44 \rightarrow 54.73$ \\
& Ours ($\alpha=16/\mu=1/\beta=10$) & 16 & 79.0 & $71.53 \rightarrow 58.35 \rightarrow 25.62$ & $75.8 \rightarrow 70.44 \rightarrow 57.28$ \\ 
& Ours ($\alpha=16/\mu=1/\beta=10$) & 1 & 72.4 & $70.1 \rightarrow 65.9 \rightarrow 46.10$ & $70.41 \rightarrow 70.31 \rightarrow 66.13$ \\  

\hline
\end{tabular}
\caption{Comparison of our approach to vanilla models against white-box attacks.}
\label{tab:adv_robustness_vanilla}
\end{table*}


A summary of the experiments is shown in Table \ref{tab:adv_robustness_vanilla}. 
Overall, for each dataset and attack type, our method shows more robustness than the vanilla model, as well as the baseline SQ @ Input. On MNIST, a significant increase in robustness is achieved by adding SQ to the input (similar to randomized smoothing), e.g.\! 15.88\% vanilla model to 76.57\% and $81.83\%$ against FGM at $\epsilon=0.3$. Interestingly, our method gave the best results with $\beta=0$ only on MNIST, potentially indicating that the representations are maximally diverse.

\textbf{$L_\infty$ robustness}: We see more interesting results on CIFAR10. For Projected Gradient Descent (PGD) on CIFAR10, the accuracy of the vanilla model is 7.81\% compared to 53.13\% of our method at $\epsilon=5/255$, more than six-fold increase in the robustness against a well-known hard attack. At $\epsilon=10/255$, we still see a six-fold improvement (vanilla model 5.54\% vs 24.79\% our method). In comparison to SQ @ Input, our method shows improved robustness, showing that the MI regularization with two SQs is better than only input SQ. Interestingly, our method gave the best results with $\beta=10$ on CIFAR10, indicating that the best robust ensemble required the feature representations between ensemble members to be forced to be diverse. The results on RESISC45 are similar to CIFAR10. Our method shows significantly higher robustness than the vanilla model. However, the gap between SQ @ Input and our method is smaller than CIFAR10, with our method being slightly ahead. 

\begin{figure}
    \centering
    \includegraphics[width=0.7\columnwidth]{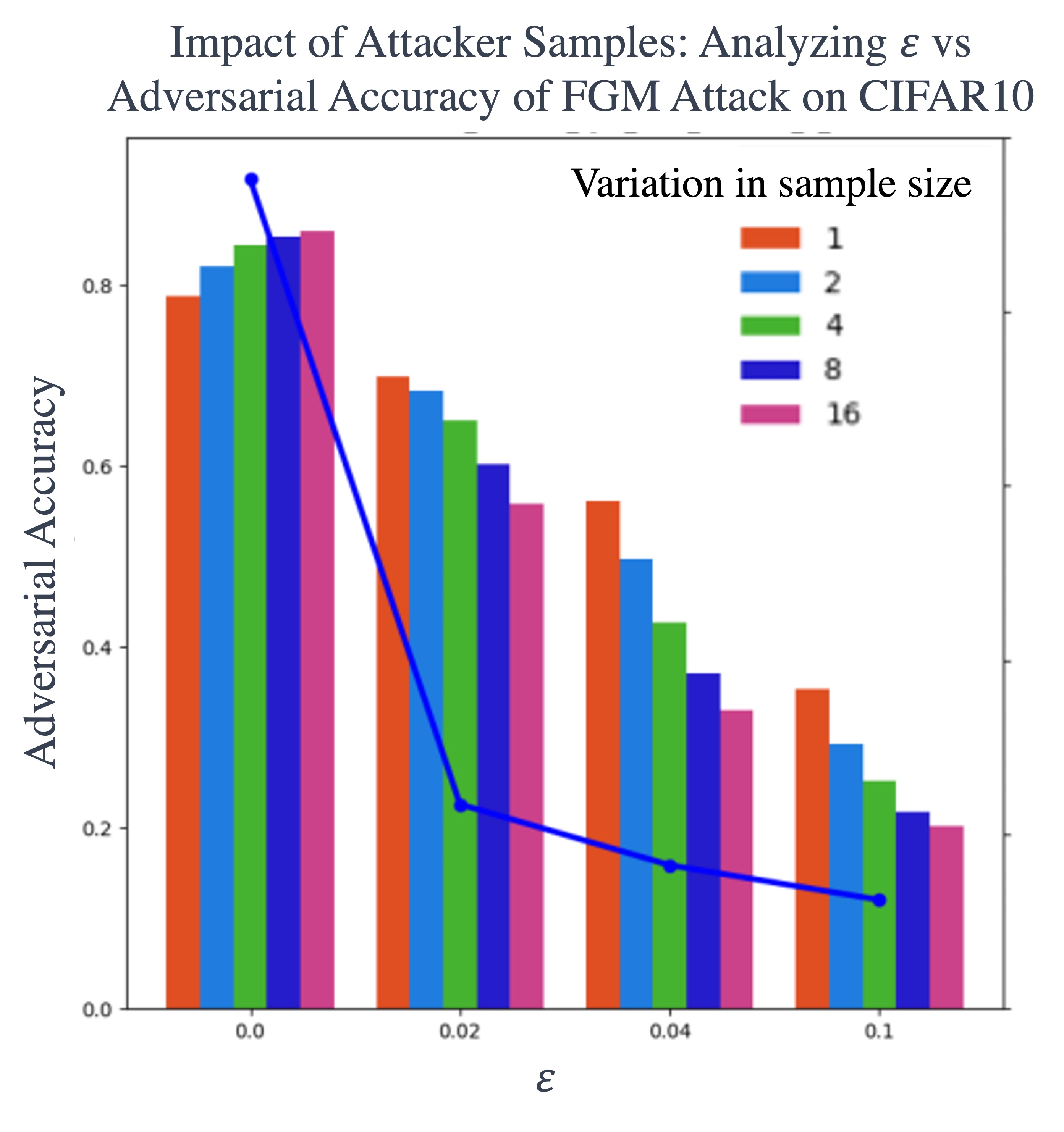} 
    \caption{Impact of the number of SQ samples used by the attacker. Blue line shows performance of the vanilla model.}
    \label{fig:variation_in_attacker_samples}
\end{figure}

\textbf{Impact of number of samples}: As expected, reducing the number of samples only degrades the ability of the attacker when using the same models as above. However, this is a best-case bound and we can never assume that only one sample will be used in practice. We further analyze its impact in Figure \ref{fig:variation_in_attacker_samples}. We observe that our method's adversarial accuracy steadily decreases as we increase the number of samples available to the attacker. Note that inference in our model uses 4-bit precision which is significantly more efficient than training the attack perturbation at full precision, especially with equal or more samples. We argue that this places an increased burden on the attacker compared to attacking the vanilla model, while still being less successful with attacks at all the tested sample sizes. 

\begin{figure}
    \centering
    \begin{subfigure}{0.49\linewidth}
        \centering
        \includegraphics[width=1.0\textwidth]{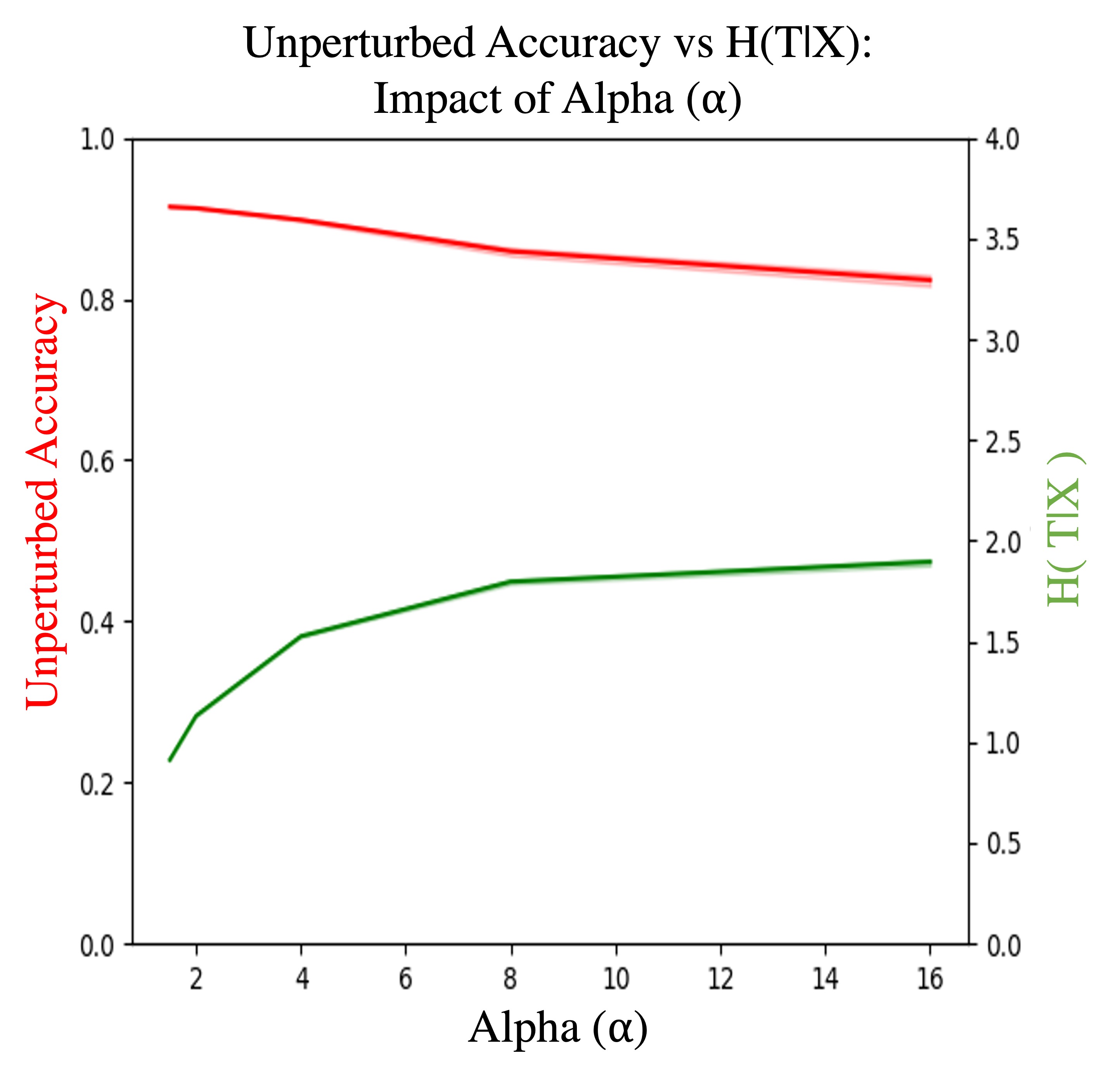}
    \end{subfigure}
    \hfill 
    \begin{subfigure}{0.49\linewidth}
        \centering
        \includegraphics[width=1.0\textwidth] {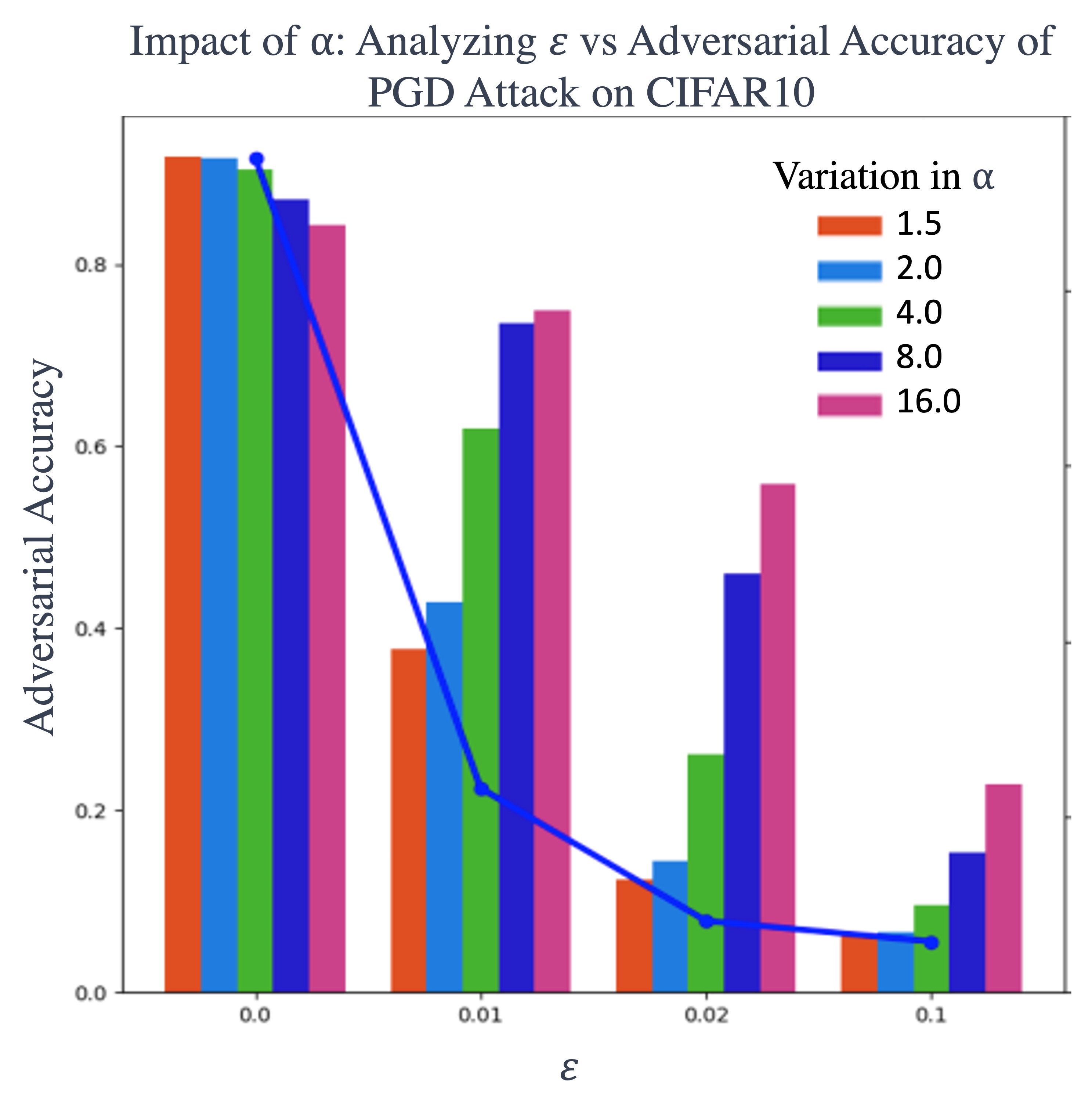}
    \end{subfigure}
    \caption{(Left) Increasing the variance $\alpha$ of SQ increases ensemble diversity $H(T|X)$. (Right) Models with higher $H(T|X)$ and higher $\alpha$ have higher adversarial robustness.}
    \label{fig:ensemble_disagreement}
\end{figure}

\textbf{Ensemble Diversity}: 
To further understand the source of improved robustness, we measured the diversity $H(\tilde{T}|\tilde{X})$ in representation across ensemble members (Fig. \ref{fig:ensemble_disagreement}). As we increase the variance $\alpha$ of SQ, we see an increasing diversity $H(T|X)$ (Left; green line). We also observed increasing prediction disagreement in the top-1 and top-3 categories (not shown). Further, we observe significantly higher adversarial robustness with increasing diversity at higher $\alpha$ (Right figure). One can see $\alpha=16$ with 16 bins as training at a very high level of noise, and the training grounded in information theory is still able to tradeoff clean accuracy for feature diversity and achieve robustness. The capacity to produce diverse DNNs is remarkable despite the constraint that the ensemble members must be stochastic quantizations of the same parent DNN. 

\textbf{Comparison to Ensemble-based and Quantization-based defenses}: 
We compared our method to the ensemble training defense ADP \cite{pmlr-v97-pang19a}  in  Table \ref{tab:adv_robustness_ensemble_adp}. ADP trains multiple DNNs at the same time, whereas our ensemble is sampled on the fly. In their paper, the authors are limited to an ensemble of three ResNet18 models on MNIST and CIFAR10. Our method shows significantly higher robustness (ADP: 52.8\% vs Ours: 91.85\% on FGM@$\epsilon$=0.2, and ADP:41.0\% vs Ours:89.53\% on PGD@$\epsilon$=.15), without significant limitations on the ensemble size. 

Next, we compared our method to \cite{sen2020empir}'s EMPIR (Table \ref{tab:adv_robustness_ensemble_empir}), where the authors leverage an ensemble of mixed precision models. Note that EMPIR authors show their results on custom MnistConv and CIFAR10Conv architectures, which we implemented and re-trained our models. We used similar strength of the attacks as that of their paper and observed improved robustness (EMPIR: 17.51\% vs Ours: 27.54\% on PGD@$\epsilon$=0.3, and EMPIR:10.69\% vs Ours:21.82\% PGD@$\epsilon$=0.1). Note that their method does not train quantized ensembles as part of the defense. 

Finally, we compare with Defensive Quantization (DQ) \cite{lin2019defensive} (Table \ref{tab:adv_robustness_ensemble_dq}). DQ focuses on Lipschitz regularization using the singular value decomposition of the weights.  We see a slight improvement but not as high as previous methods when our method is applied to their WideResNet18 architecture and evaluated on CIFAR10 against FGM attack. We were unable to reproduce their method on other datasets and attacks. 


\begin{table*}[ht]
\centering
\begin{tabular}{|c|c|c|c|c|}
\hline
Dataset & Ensemble Methods & Unperturbed & FGM (0.1 $\rightarrow$ 0.2) & PGD (0.1  $\rightarrow$  0.15) \\
\hline
\multirow{4}{*}{MNIST} & ADP-baseline \cite{pmlr-v97-pang19a} & NA & 78.3$\rightarrow$21.5 & 50.7$\rightarrow$6.3 \\ 
& ADP \cite{pmlr-v97-pang19a} & NA & 96.3$\rightarrow$52.8 & 82.8$\rightarrow$41.0 \\ 
& Ours (4/1/10/16) & 99.53 & 98.41$\rightarrow$91.85 & 97.09$\rightarrow$89.53 \\ 
& Ours (4/1/10/1) & 99.53 & \textbf{98.43$\rightarrow$92.31} & \textbf{97.38$\rightarrow$93.08} \\ 
\hline
 &  &  & FGM (0.02$\rightarrow$0.04) & PGD (0.01$\rightarrow$0.02) \\
\hline

\multirow{4}{*}{CIFAR10} & ADP-baseline \cite{pmlr-v97-pang19a} & NA & 36.5$\rightarrow$19.4 & 23.6$\rightarrow$6.6 \\ 
& ADP \cite{pmlr-v97-pang19a} & NA & 61.7$\rightarrow$46.2 & 48.4$\rightarrow$30.4 \\ 
& Ours (16/1/10/16) & 85.93 & 55.79$\rightarrow$32.99 & 74.5$\rightarrow$53.13 \\ 
& Ours (16/1/10/1) & 78.80 & \textbf{69.93$\rightarrow$56.14} & \textbf{75.93$\rightarrow$70.1} \\ 

\hline
\end{tabular}
\caption{Robustness comparison to ADP, Model: ResNet-18}
\label{tab:adv_robustness_ensemble_adp}
\end{table*}



\begin{table*}[ht]
\centering
\begin{tabular}{|c|c|c|c|c|}
\hline
Dataset & Ensemble Methods & Unperturbed & FGM (0.3) & PGD (0.3) \\
\hline
\multirow{4}{*}{MNIST} & EMPIR-baseline \cite{sen2020empir} & 98.87 & 14.32 & 0.77 \\ 
& EMPIR \cite{sen2020empir} & 98.89 & 67.06 & 17.51 \\ 
& Ours (16/1/5/16) & 99.33 & 84.63 & 38.33 \\ 
& Ours (16/1/5/1) & 99.09 & 82.56 & 27.94 \\ 
\hline
 &  &  & FGM (0.1) & PGD (0.1) \\
\hline

\multirow{4}{*}{CIFAR10} & EMPIR-baseline \cite{sen2020empir} & 74.54 & 10.28 & 10.69 \\ 
& EMPIR \cite{sen2020empir} & 72.56 & 20.45 & 13.55 \\ 
& Ours (16/0/5/16) & 75.87 & 15.1 & 21.82 \\ 
& Ours (16/0/5/1) & 65.27 & \textbf{28.77} & \textbf{50.1} \\ 

\hline
\end{tabular}
\caption{Robustness comparison to EMPIR, Model MNISTConv for MNIST and CIFARConv for CIFAR10}
\label{tab:adv_robustness_ensemble_empir}
\end{table*}



\begin{table}[ht]
\centering
\footnotesize
\begin{tabular}{|c|c|c|c|c|}
\hline
Dataset & Quantization Methods & Unperturbed & FGM (8/255)  \\
\hline
\multirow{4}{*}{CIFAR10} & DQ-Vanilla \cite{lin2019defensive} & 93.9 & 19.0 \\ 
& DQ-4bits \cite{lin2019defensive} & 95.8 & 49.8 \\ 
& Ours (4/1/10/16) & 89.41 & 50.1 \\ 
& Ours (4/1/10/1) & 88.20 & 61.4 \\ 

\hline
\end{tabular}
\caption{Robustness comparison to DQ, Model: WideResNet}
\label{tab:adv_robustness_ensemble_dq}
\end{table}

\begin{figure*}
    \centering
    \hfill
    \begin{minipage}[t]{0.3\linewidth}
    \centering
    \includegraphics[width=0.8\linewidth]{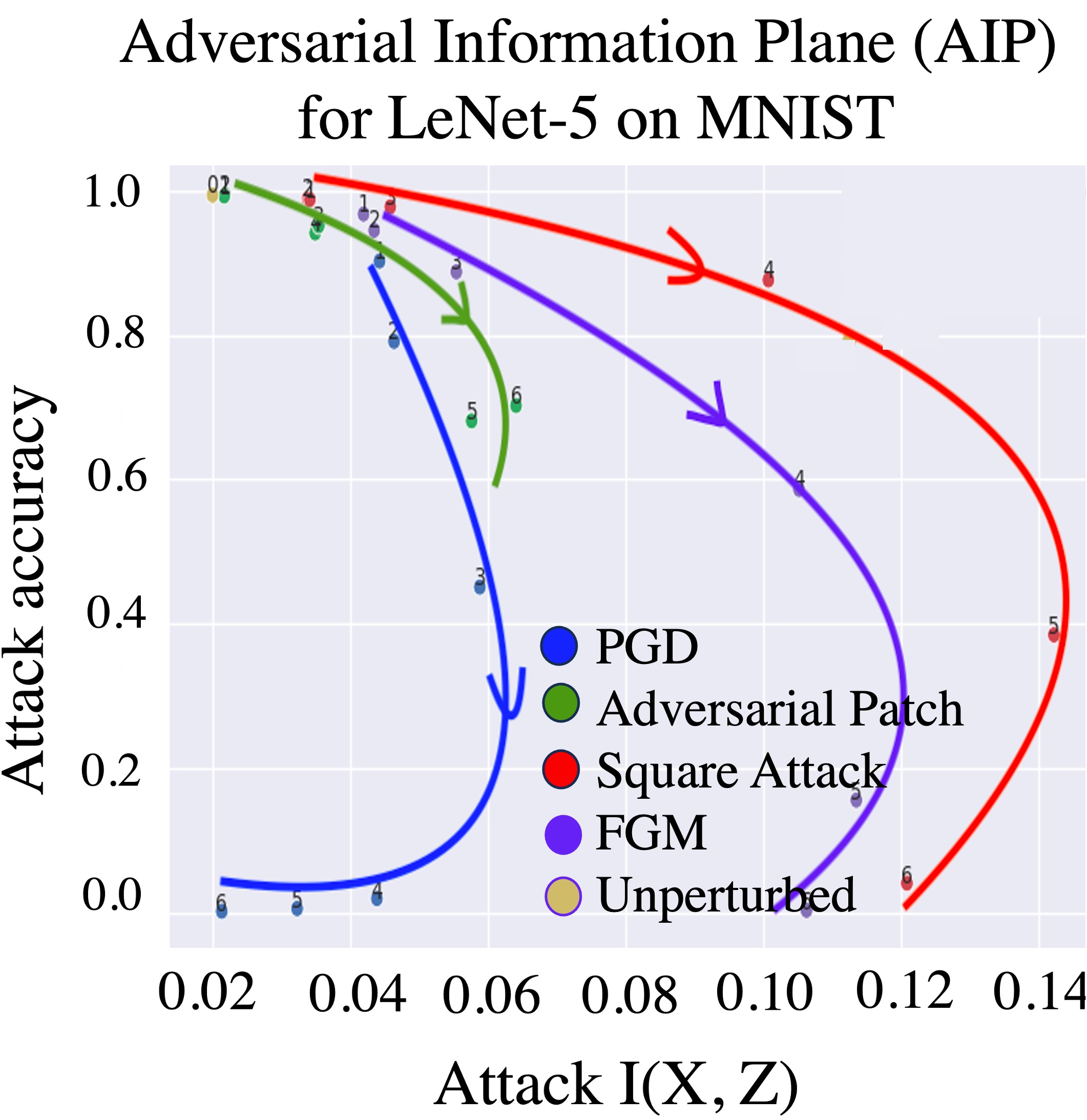}
    \end{minipage}
    \hfill
    \begin{minipage}[t]{0.3\linewidth}
    \centering
    \includegraphics[width=0.8\linewidth]{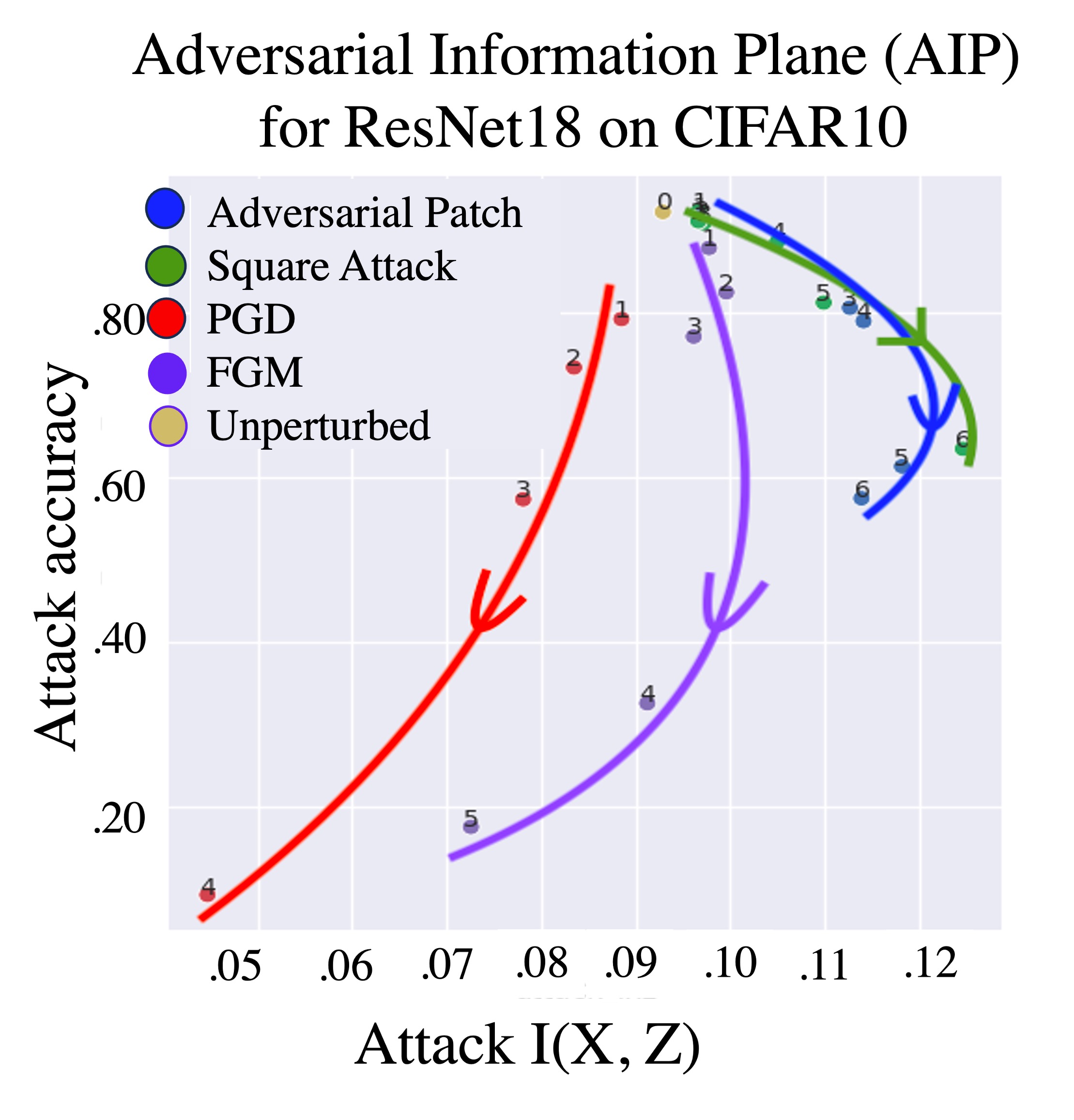}
    \end{minipage}
    \hfill
    \begin{minipage}[t]{0.3\linewidth}
    \centering
    \includegraphics[width=0.8\linewidth]{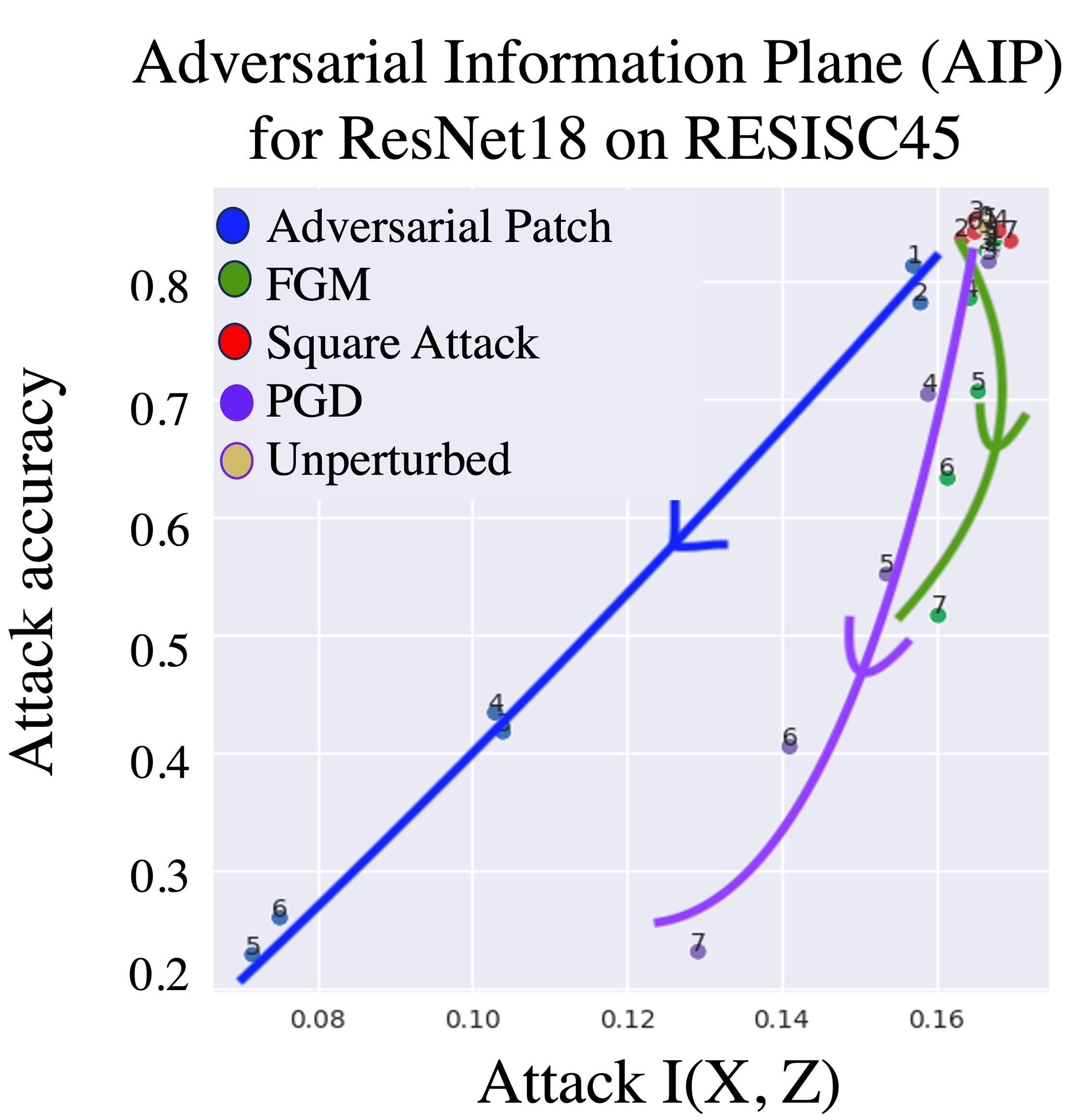}
    \end{minipage}
    \hfill
    \caption{Adversarial Information Plane (AIP) to understand the vulnerabilities of different DNNs and datasets. The attack strength is annotated on each data point, e.g., for FGM 1-6 corresponds to $\epsilon$ from 0.05 to 0.5 for MNIST.}
    \label{fig:AttackProfile}
\end{figure*}




\textbf{Adversarial Information Plane (AIP) visualization}: 
Our method does not assume an explicit threat model of attacks. Information bottleneck theory \cite{tishby2000information} suggests that MI-like statistics might shed general insight into DNN learning and performance. Admittedly, our MI is significantly different than theirs, but nonetheless, we attempt a unified look at different attack types. AIP plots (Fig. \ref{fig:AttackProfile}) illustrate changes in mutual information (MI) values and accuracy with increasing attack strength. These MI are generated by our method when evaluated on attacks of four different types: FGM, PDG, Square Attack, and Patch attack. Each attack type is evaluated at multiple strengths and the direction of the curve indicates increasing attack strength. 

In general, for all attacks, we observe the MI $I(\tilde{X},\tilde{T})$ increases at first as the attack strength increases, then decreases with a further increase in attack strength, as the accuracy (y-axis) constantly decreases with attack strength. This is because ensemble members will produce similar representations when an extreme amount of noise is added that washes out the effect of quantization noise in a single member. We observe that the AIP visualization allows extrapolating the vulnerability to different attack strengths without testing against all of them. 

We observe that there are some adversarial attacks that decrease accuracy without changing MI significantly. The lines that point vertically down are particularly insidious, as they can be difficult to distinguish from clean data based solely on MI. For example, the range of MI values induced by PGD on MNIST is smaller than the range of MI values under FGM. We know from the literature that PGD attacks are harder to detect than FGM attacks, correlated with the MI values shown in the AIP. On RESISC45, we see that larger-size patch attacks more drastically reduce MI, which is also correlated with relative ease of attack detection. In contrast, FGM and PGD do not change the MI drastically in these larger images. These plots allow for quantifying the relative difficulty of different attacks on a single plot and yield insights about attack detection that seem correlated with practice.

\begin{figure}
    \centering
    \includegraphics[width=0.7\linewidth]{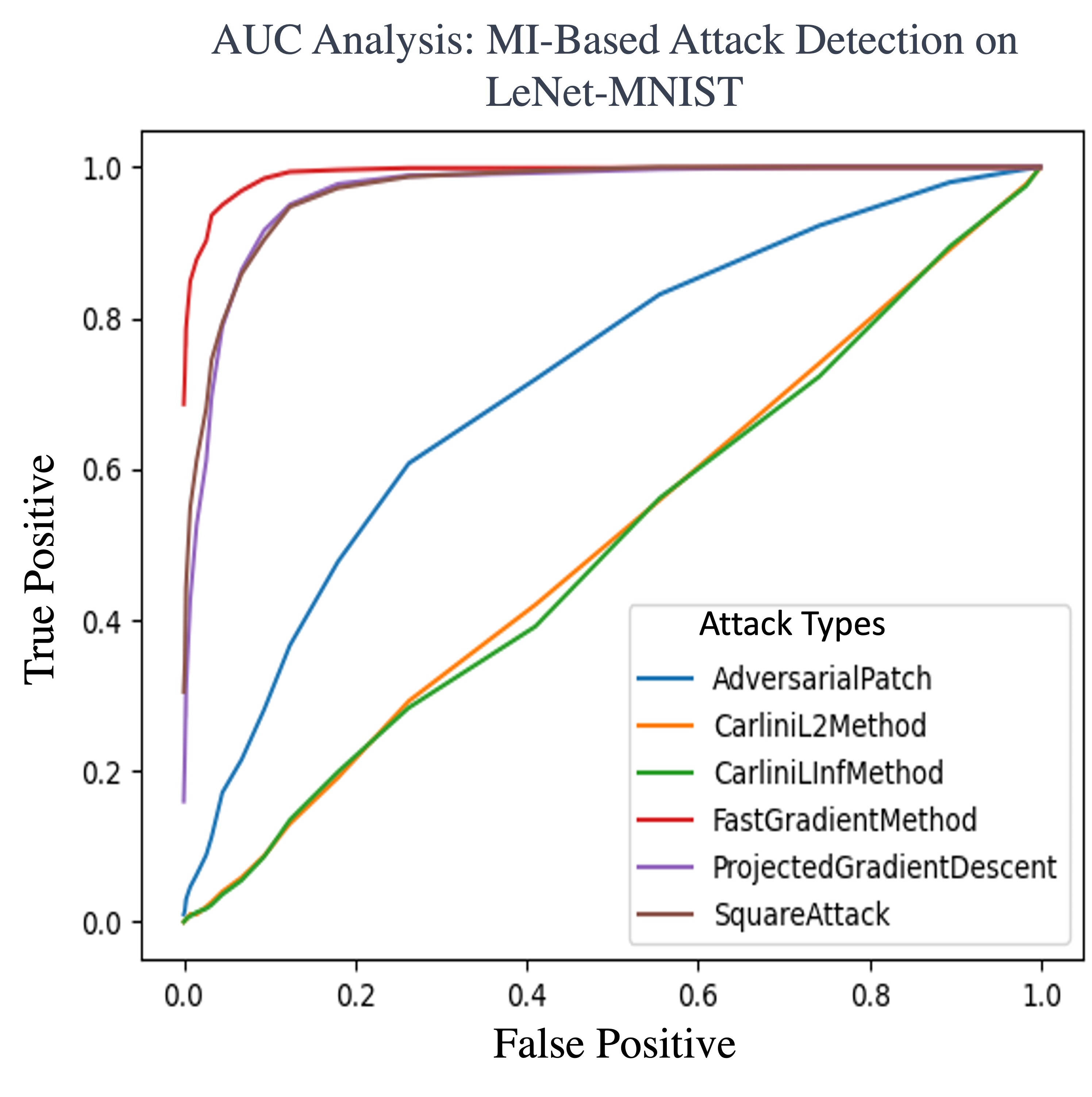} 
    \caption{ROC curve: attack detection on LeNet-MNIST ($\alpha$ = 16)}
    \label{fig:lenet_detection}
\end{figure}

\textbf{Attack detection using MI}: The AIP visualization suggests that only some attacks can be detected via the MI of the ensemble. We created an MI-based attack detector to confirm the relative hardness suggested by the AIP.  The detector works by calculating a threshold average MI from clean data and checking the MI of test examples against the threshold. Figure~\ref{fig:lenet_detection} shows ROC curves for attack detection. The ROC curves are generated by calculating the average MI of unperturbed data and predicting an attack if  MI is greater than this average by some threshold and sweeping through all thresholds. We observe that our detector works well for L$\infty$ attacks and patch attacks, but doesn't work at all for Carlini\&Wagner attack. 

\begin{figure}
    \begin{subfigure}{0.45\linewidth}
        \centering
        \includegraphics[width=1.0\linewidth]{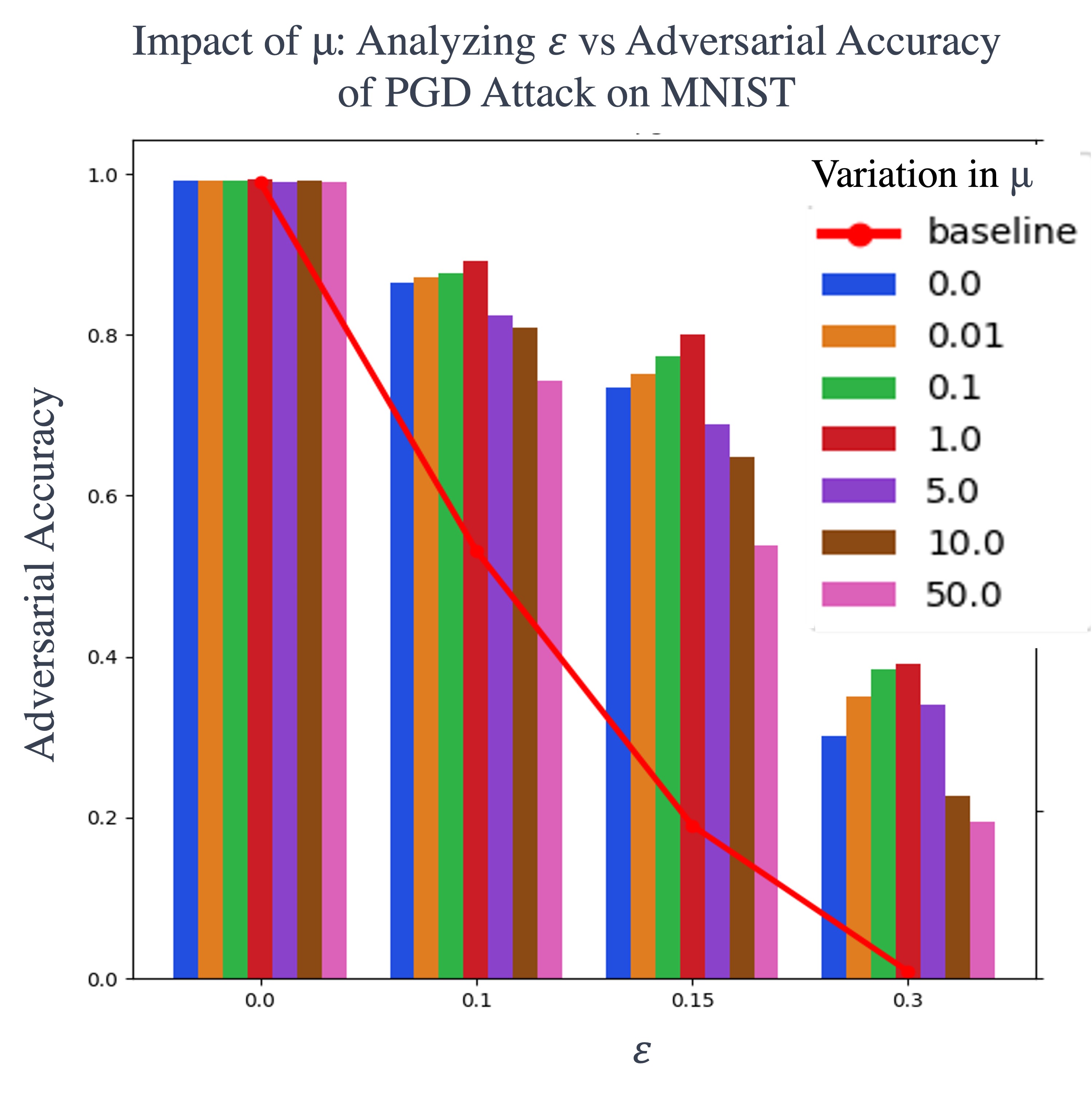}
    \end{subfigure}
    \hfill 
    \begin{subfigure}{0.45\linewidth}
        \centering
        \includegraphics[width=1.0\linewidth]{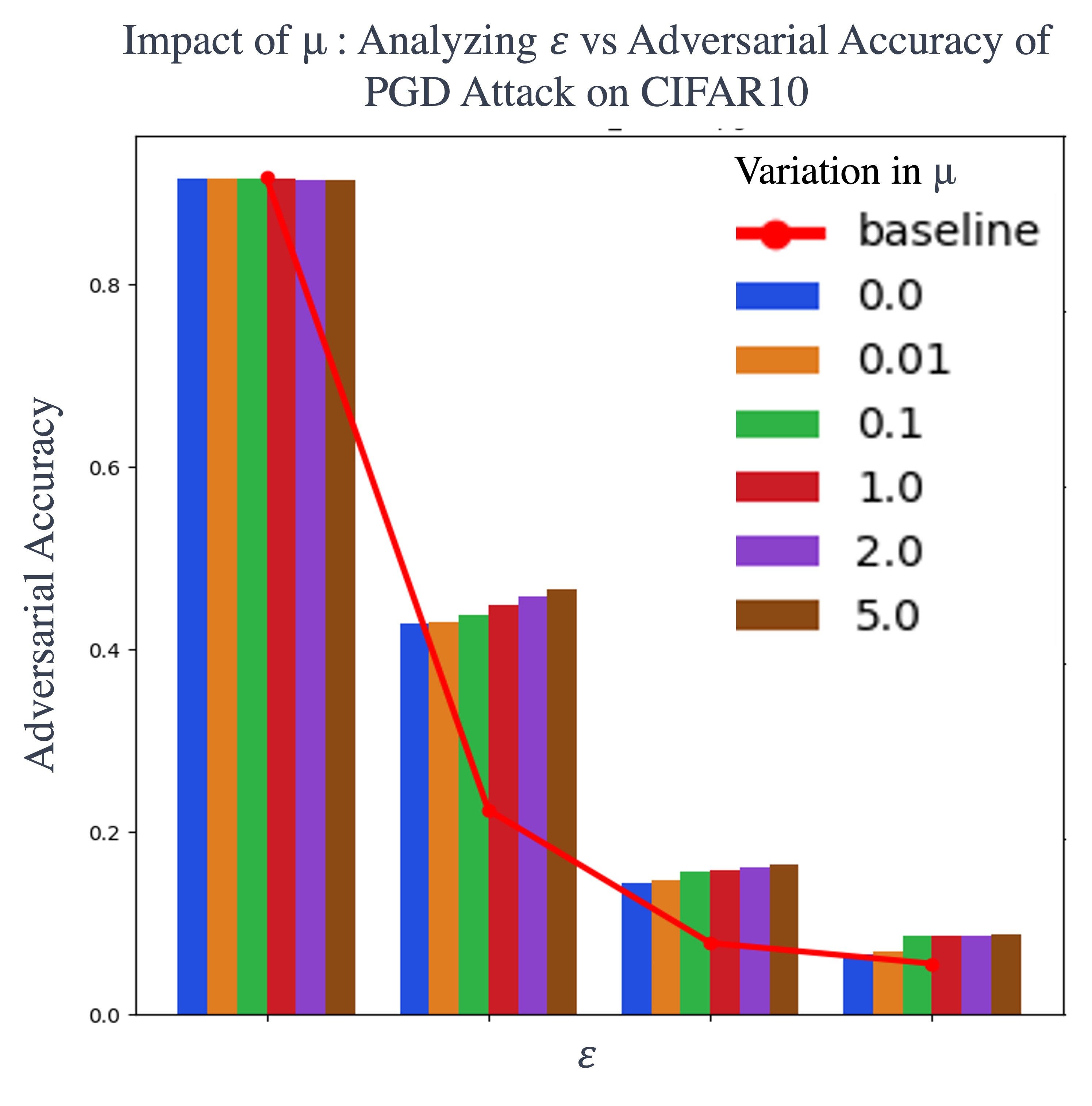}
    \end{subfigure}
    \caption{Ablation on $\mu$ for noise regularization (in Eq. \ref{eq:qed_loss_final}): improves robustness to PGD attack. Other attacks are not shown but the results were similar.}
        \label{fig:qed_lipschitz_reg}
\end{figure}

\textbf{Ablating effect of regularization ($\mu$)}: 
Figure \ref{fig:qed_lipschitz_reg} shows varying $\mu$ during training while keeping $\beta$ and $\alpha$ fixed. Evaluation is performed using the PGD attack on LeNet-MNIST and ResNet18-CIFAR10. We observe that as we increase $\mu$, the Lipschitz regularization, our model shows increased robustness up to a certain value of $\mu$. After that point, the model collapses (for $\mu=5$ in MNIST) and performs at chance level, indicating that the minimum and maximum of the sampled feature have collapsed.

\begin{figure}
    \begin{subfigure}{0.45\linewidth}
        \centering
        \includegraphics[width=1.0\linewidth]{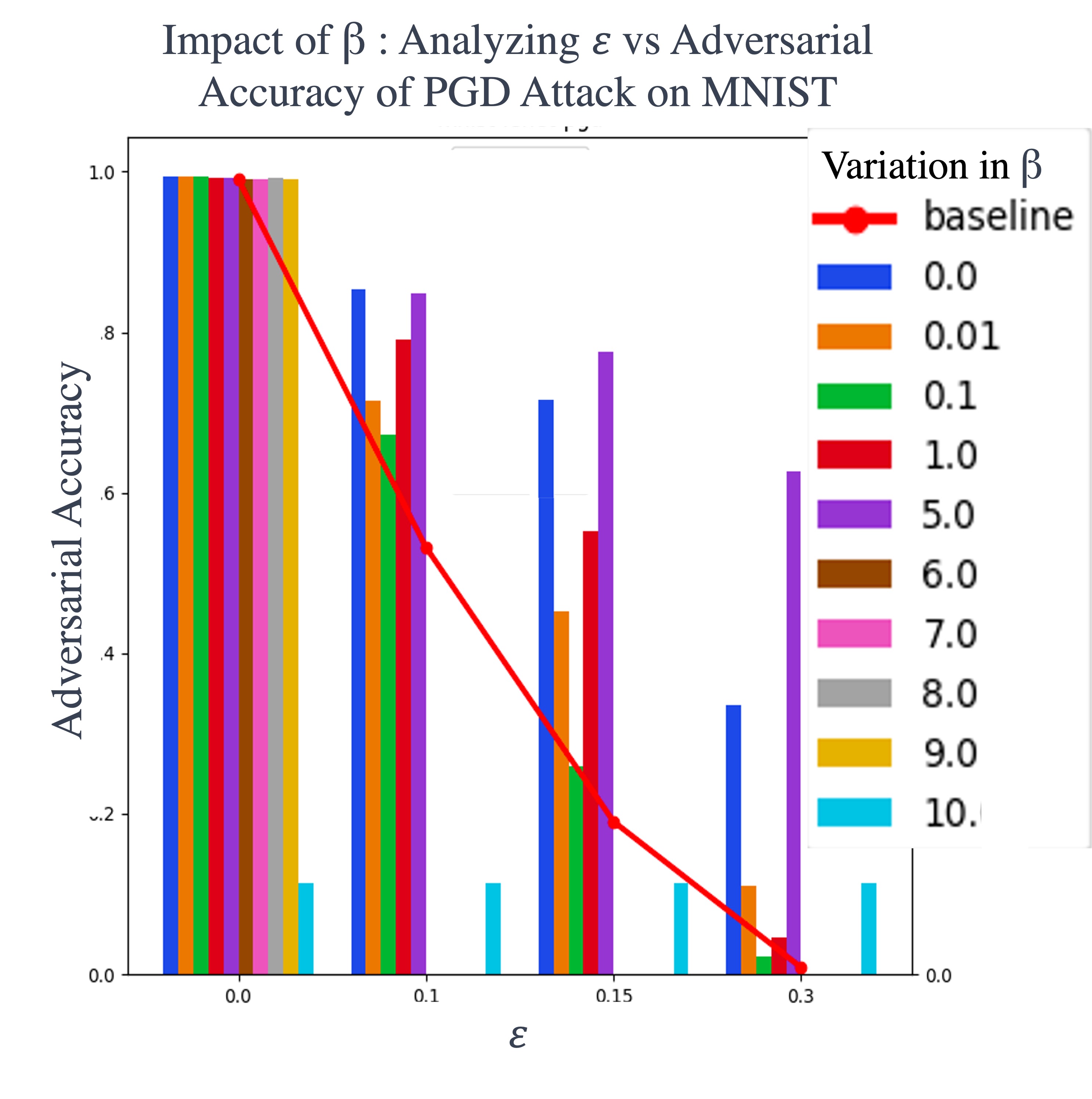}
    \end{subfigure}
    \hfill 
    \begin{subfigure}{0.45\linewidth}
        \centering
        \includegraphics[width=1.0\linewidth]{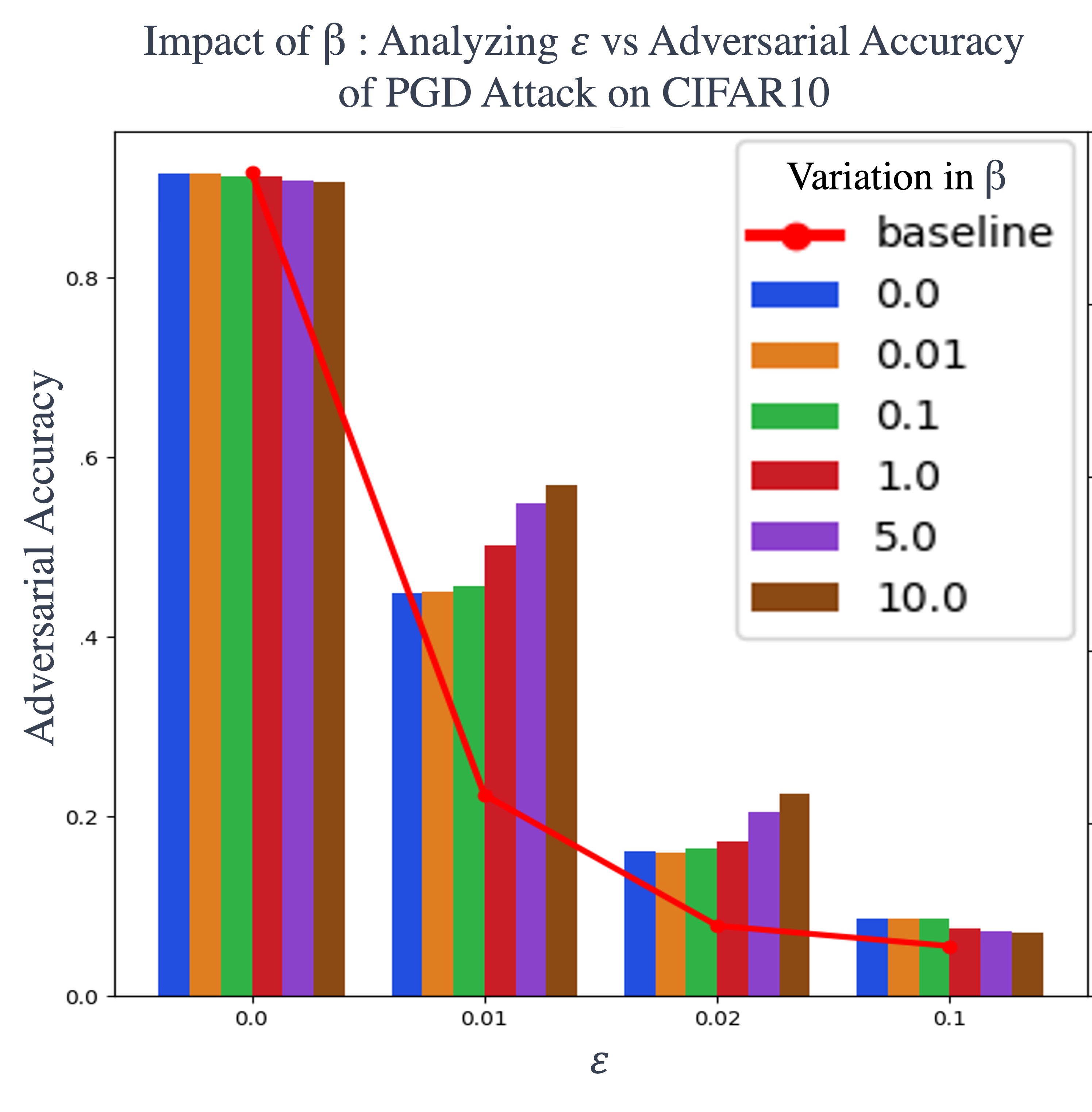}
    \end{subfigure}
    \caption{Ablation on $\beta$ for MI regularization (in Eq. \ref{eq:qed_loss_final}): improves robustness to PGD attack. Other attacks are not shown but the results were similar.}
    \label{fig:qed_mi_reg}
\end{figure}

\textbf{Ablating effect of MI Regularization ($\beta$)}: 
Figure \ref{fig:qed_mi_reg} shows varying $\beta$ during training while keeping $\mu$ and $\alpha$ fixed. Evaluation is done using the PGD attack on Lenet-MNIST and ResNet18-CIFAR10. Larger $\beta$ values lead to higher robustness in CIFAR10. The ablation on MNIST did not result in the same smooth trend. 

%

\section{Discussion and Related Work}

Despite active research on adversarial attacks and defenses, the robustness of deep neural networks to adversarial attacks remains an unsolved problem. Adversarial robustness of quantized DNNs is under-explored and there is a need for methods that can mitigate the drop in robustness due to quantization. Our approach is to train an ensemble of different quantized DNNs from a single DNN. Our approach is fast and efficient, while the training is grounded in information-theory leading to significant gains in robustness to $L_\infty$ attacks in a considerably fair evaluation. The mutual information (MI) estimated within the ensemble provides an additional tool to quantify the vulnerabilities against different attacks in a unified manner. Future work could consider integrating this approach with adversarial training, e.g.\! to intelligently sample the quantized images used for training. 

We conclude with a brief (and by no means exhaustive) review of defenses in the literature that have inspired the development of our defense. The core idea behind ensemble defenses is to combine strength from multiple models to enhance overall robustness collectively. \cite{strauss2017ensemble} explores ensemble methods by initializing models with different weights to improve robustness, whereas \cite{pang2019improving} adds regularization to train diverse sets of ensemble models. \cite{sen2020empir} ensembled quantized DNN models with different numerical precision that prompted our initial hypothesis. \cite{tramr2018ensemble} introduces ensemble adversarial training, a technique that augments training data with perturbations transferred from other models. \cite{abbasi2017robustness} propose to use an ensemble of diverse specialists based on a confusion matrix. \cite{he2017adversarial} combines multiple weak defenses and concludes that it is insufficient to create a more robust defense. \cite{liu2018towards} introduced a defense algorithm called Random Self Ensemble (RSE) that enhances the security of neural networks by adding random noise to layers and ensemble predictions, effectively defending against gradient-based attacks while maintaining predictive capability without significant memory overhead. 

On the other hand, quantization aims to reduce the precision of Neural network weights and activation. \cite{gholami2021survey}'s survey talks about recent advances and techniques used to quantize DNNs in training and post-training. \cite{zhou2020neural} use rate-distortion theory to determine the most significant bits in the feature representation and related to our information-theoretic approach. \cite{yang2019quantization} presents a low-bit differentiable quantization function, which can be learned in a lossless and end-to-end manner for any weight and activation. 
\cite{lin2019defensive} introduced defensive quantization by controlling the Lipschitz constant of the quantized DNN. 
\cite{rakin2018defend} proposed adaptive quantization techniques to quantize activation functions. These prior works do not consider SQ or creating an ensemble of quantized DNNs that can also be trained from scratch. 

\textbf{Adversarial attack detection.}
Attack detection refers to identifying adversarial attacks on the DNNs. \cite{bagnall2017training} proposed an ensemble method for detecting and classifying adversarial attacks. \cite{feinman2017detecting} uses Bayesian uncertainty estimates in the feature spaces to find artifacts in learned space. \cite{metzen2017detecting} trains a small detector sub-network for binary classification problems to distinguish between clean and adversarial data. \cite{wang2021multi} proposed a MultiExpert Adversarial Attack Detection (MEAAD) approach by checking context inconsistency, suitable for any DNN-based ReID systems. \cite{xiang2021patchguard++} uses a small receptive field of CNNs to detect attacks in feature space. \cite{smith2018understanding}. \cite{lee2018simple} proposes a unified framework to detect OOD and adversarial samples using Gaussian discriminant analysis and applies it to any pre-trained softmax neural classifier. \cite{carlini2017adversarial} evaluates various attack detection algorithms and provides guidelines for evaluation. These methods do not use a sufficient statistic like MI to detect attacks uniformly across threat models.

\section*{Acknowledgement}
This material is based upon work supported by the Defense Advanced Research Projects Agency (DARPA) under Agreement No. HR00111990078. Any opinions, findings and conclusions or recommendations expressed in this material are those of the authors and do not necessarily reflect the views of the Defense Advanced Research Projects Agency (DARPA). 

{\small
\bibliographystyle{ieee_fullname}
\bibliography{egbib}

\begin{thebibliography}{10}\itemsep=-1pt

\bibitem{abbasi2017robustness}
Mahdieh Abbasi and Christian Gagn{\'e}.
\newblock Robustness to adversarial examples through an ensemble of specialists.
\newblock {\em arXiv preprint arXiv:1702.06856}, 2017.

\bibitem{alemi2016deep}
Alexander~A Alemi, Ian Fischer, Joshua~V Dillon, and Kevin Murphy.
\newblock Deep variational information bottleneck.
\newblock {\em arXiv preprint arXiv:1612.00410}, 2016.

\bibitem{ACFH2020square}
Maksym Andriushchenko, Francesco Croce, Nicolas Flammarion, and Matthias Hein.
\newblock Square attack: a query-efficient black-box adversarial attack via random search.
\newblock {\em ECCV}, 2020.

\bibitem{athalye2018synthesizing}
Anish Athalye, Logan Engstrom, Andrew Ilyas, and Kevin Kwok.
\newblock Synthesizing robust adversarial examples.
\newblock In {\em International conference on machine learning}, pages 284--293. PMLR, 2018.

\bibitem{bagnall2017training}
Alexander Bagnall, Razvan Bunescu, and Gordon Stewart.
\newblock Training ensembles to detect adversarial examples.
\newblock {\em arXiv preprint arXiv:1712.04006}, 2017.

\bibitem{banner2018scalable}
Ron Banner, Itay Hubara, Elad Hoffer, and Daniel Soudry.
\newblock Scalable methods for 8-bit training of neural networks.
\newblock {\em Advances in neural information processing systems}, 31, 2018.

\bibitem{carlini2017adversarial}
Nicholas Carlini and David Wagner.
\newblock Adversarial examples are not easily detected: Bypassing ten detection methods.
\newblock In {\em Proceedings of the 10th ACM workshop on artificial intelligence and security}, pages 3--14, 2017.

\bibitem{Cheng_2017}
Gong Cheng, Junwei Han, and Xiaoqiang Lu.
\newblock Remote sensing image scene classification: Benchmark and state of the art.
\newblock {\em Proceedings of the IEEE}, 105(10):1865--1883, Oct 2017.

\bibitem{cohen2019certified}
Jeremy Cohen, Elan Rosenfeld, and Zico Kolter.
\newblock Certified adversarial robustness via randomized smoothing.
\newblock In {\em International Conference on Machine Learning}, pages 1310--1320. PMLR, 2019.

\bibitem{courbariaux2015binaryconnect}
Matthieu Courbariaux, Yoshua Bengio, and Jean-Pierre David.
\newblock Binaryconnect: Training deep neural networks with binary weights during propagations.
\newblock {\em Advances in neural information processing systems}, 28, 2015.

\bibitem{deng2012mnist}
Li Deng.
\newblock The mnist database of handwritten digit images for machine learning research.
\newblock {\em IEEE Signal Processing Magazine}, 29(6):141--142, 2012.

\bibitem{feinman2017detecting}
Reuben Feinman, Ryan~R Curtin, Saurabh Shintre, and Andrew~B Gardner.
\newblock Detecting adversarial samples from artifacts.
\newblock {\em arXiv preprint arXiv:1703.00410}, 2017.

\bibitem{finlay2018lipschitz}
Chris Finlay, Jeff Calder, Bilal Abbasi, and Adam Oberman.
\newblock Lipschitz regularized deep neural networks generalize and are adversarially robust.
\newblock {\em arXiv preprint arXiv:1808.09540}, 2018.

\bibitem{gholami2021survey}
Amir Gholami, Sehoon Kim, Zhen Dong, Zhewei Yao, Michael~W Mahoney, and Kurt Keutzer.
\newblock A survey of quantization methods for efficient neural network inference.
\newblock {\em arXiv preprint arXiv:2103.13630}, 2021.

\bibitem{goodfellow2014explaining}
Ian~J Goodfellow, Jonathon Shlens, and Christian Szegedy.
\newblock Explaining and harnessing adversarial examples.
\newblock {\em arXiv preprint arXiv:1412.6572}, 2014.

\bibitem{guo2018survey}
Yunhui Guo.
\newblock A survey on methods and theories of quantized neural networks.
\newblock {\em arXiv preprint arXiv:1808.04752}, 2018.

\bibitem{He_2016_CVPR}
Kaiming He, Xiangyu Zhang, Shaoqing Ren, and Jian Sun.
\newblock Deep residual learning for image recognition.
\newblock In {\em Proceedings of the IEEE Conference on Computer Vision and Pattern Recognition (CVPR)}, June 2016.

\bibitem{he2017adversarial}
Warren He, James Wei, Xinyun Chen, Nicholas Carlini, and Dawn Song.
\newblock Adversarial example defense: Ensembles of weak defenses are not strong.
\newblock In {\em 11th USENIX workshop on offensive technologies (WOOT 17)}, 2017.

\bibitem{jang2016categorical}
Eric Jang, Shixiang Gu, and Ben Poole.
\newblock Categorical reparameterization with gumbel-softmax.
\newblock {\em arXiv preprint arXiv:1611.01144}, 2016.

\bibitem{Krizhevsky09learningmultiple}
Alex Krizhevsky.
\newblock Learning multiple layers of features from tiny images.
\newblock Technical report, University of Toronto, 2009.

\bibitem{lecun1998gradient}
Yann LeCun, L{\'e}on Bottou, Yoshua Bengio, and Patrick Haffner.
\newblock Gradient-based learning applied to document recognition.
\newblock {\em Proceedings of the IEEE}, 86(11):2278--2324, 1998.

\bibitem{lee2018simple}
Kimin Lee, Kibok Lee, Honglak Lee, and Jinwoo Shin.
\newblock A simple unified framework for detecting out-of-distribution samples and adversarial attacks.
\newblock {\em Advances in neural information processing systems}, 31, 2018.

\bibitem{li2016ternary}
Fengfu Li, Bo Zhang, and Bin Liu.
\newblock Ternary weight networks.
\newblock {\em arXiv preprint arXiv:1605.04711}, 2016.

\bibitem{lin2019defensive}
Ji Lin, Chuang Gan, and Song Han.
\newblock Defensive quantization: When efficiency meets robustness.
\newblock {\em arXiv preprint arXiv:1904.08444}, 2019.

\bibitem{liu2018towards}
Xuanqing Liu, Minhao Cheng, Huan Zhang, and Cho-Jui Hsieh.
\newblock Towards robust neural networks via random self-ensemble.
\newblock In {\em Proceedings of the European Conference on Computer Vision (ECCV)}, pages 369--385, 2018.

\bibitem{madry2017towards}
Aleksander Madry, Aleksandar Makelov, Ludwig Schmidt, Dimitris Tsipras, and Adrian Vladu.
\newblock Towards deep learning models resistant to adversarial attacks.
\newblock {\em arXiv preprint arXiv:1706.06083}, 2017.

\bibitem{metzen2017detecting}
Jan~Hendrik Metzen, Tim Genewein, Volker Fischer, and Bastian Bischoff.
\newblock On detecting adversarial perturbations.
\newblock {\em arXiv preprint arXiv:1702.04267}, 2017.

\bibitem{pang2019improving}
Tianyu Pang, Kun Xu, Chao Du, Ning Chen, and Jun Zhu.
\newblock Improving adversarial robustness via promoting ensemble diversity.
\newblock In {\em International Conference on Machine Learning}, pages 4970--4979. PMLR, 2019.

\bibitem{pmlr-v97-pang19a}
Tianyu Pang, Kun Xu, Chao Du, Ning Chen, and Jun Zhu.
\newblock Improving adversarial robustness via promoting ensemble diversity.
\newblock In Kamalika Chaudhuri and Ruslan Salakhutdinov, editors, {\em Proceedings of the 36th International Conference on Machine Learning}, volume~97 of {\em Proceedings of Machine Learning Research}, pages 4970--4979. PMLR, 09--15 Jun 2019.

\bibitem{parajuli2018generalized}
Samyak Parajuli, Aswin Raghavan, and Sek Chai.
\newblock Generalized ternary connect: End-to-end learning and compression of multiplication-free deep neural networks.
\newblock {\em arXiv preprint arXiv:1811.04985}, 2018.

\bibitem{pensia2020extracting}
Ankit Pensia, Varun Jog, and Po-Ling Loh.
\newblock Extracting robust and accurate features via a robust information bottleneck.
\newblock {\em IEEE Journal on Selected Areas in Information Theory}, 1(1):131--144, 2020.

\bibitem{rakin2018defend}
Adnan~Siraj Rakin, Jinfeng Yi, Boqing Gong, and Deliang Fan.
\newblock Defend deep neural networks against adversarial examples via fixed and dynamic quantized activation functions.
\newblock {\em arXiv preprint arXiv:1807.06714}, 2018.

\bibitem{sen2020empir}
Sanchari Sen, Balaraman Ravindran, and Anand Raghunathan.
\newblock Empir: Ensembles of mixed precision deep networks for increased robustness against adversarial attacks.
\newblock {\em arXiv preprint arXiv:2004.10162}, 2020.

\bibitem{shwartz2020information}
Ravid Shwartz-Ziv and Alexander~A Alemi.
\newblock Information in infinite ensembles of infinitely-wide neural networks.
\newblock In {\em Symposium on Advances in Approximate Bayesian Inference}, pages 1--17. PMLR, 2020.

\bibitem{smith2018understanding}
Lewis Smith and Yarin Gal.
\newblock Understanding measures of uncertainty for adversarial example detection.
\newblock {\em arXiv preprint arXiv:1803.08533}, 2018.

\bibitem{strauss2017ensemble}
Thilo Strauss, Markus Hanselmann, Andrej Junginger, and Holger Ulmer.
\newblock Ensemble methods as a defense to adversarial perturbations against deep neural networks.
\newblock {\em arXiv preprint arXiv:1709.03423}, 2017.

\bibitem{tishby2000information}
Naftali Tishby, Fernando~C Pereira, and William Bialek.
\newblock The information bottleneck method.
\newblock {\em arXiv preprint physics/0004057}, 2000.

\bibitem{tramr2018ensemble}
Florian Tramèr, Alexey Kurakin, Nicolas Papernot, Ian Goodfellow, Dan Boneh, and Patrick McDaniel.
\newblock Ensemble adversarial training: Attacks and defenses.
\newblock In {\em International Conference on Learning Representations}, 2018.

\bibitem{wang2021multi}
Xueping Wang, Shasha Li, Min Liu, Yaonan Wang, and Amit~K Roy-Chowdhury.
\newblock Multi-expert adversarial attack detection in person re-identification using context inconsistency.
\newblock In {\em Proceedings of the IEEE/CVF International Conference on Computer Vision}, pages 15097--15107, 2021.

\bibitem{xiang2021patchguard++}
Chong Xiang and Prateek Mittal.
\newblock Patchguard++: Efficient provable attack detection against adversarial patches.
\newblock {\em arXiv preprint arXiv:2104.12609}, 2021.

\bibitem{yang2019quantization}
Jiwei Yang, Xu Shen, Jun Xing, Xinmei Tian, Houqiang Li, Bing Deng, Jianqiang Huang, and Xian-sheng Hua.
\newblock Quantization networks.
\newblock In {\em Proceedings of the IEEE/CVF Conference on Computer Vision and Pattern Recognition}, pages 7308--7316, 2019.

\bibitem{zhou2020neural}
Xichuan Zhou, Kui Liu, Cong Shi, Haijun Liu, and Ji Liu.
\newblock Neural network activation quantization with bitwise information bottlenecks.
\newblock {\em arXiv preprint arXiv:2006.05210}, 2020.

\end{thebibliography}
}

\end{document}